\documentclass[preprint]{elsarticle}

\makeatletter
\def\ps@pprintTitle{%
 \let\@oddhead\@empty
 \let\@evenhead\@empty
 \def\@oddfoot{}%
 \let\@evenfoot\@oddfoot}
\makeatother

\usepackage{amsmath,amssymb,amsthm}
\usepackage{amsthm}
\usepackage{verbatim}
\usepackage{upgreek}
\usepackage{color}
\usepackage{xcolor}
\usepackage{algorithmic}
\usepackage[ruled, lined, linesnumbered]{algorithm2e}
\usepackage{caption}
\usepackage{float}
\usepackage{multicol}
\usepackage{multirow}
\usepackage{pdflscape}

\usepackage{array}
\newcolumntype{L}{>{\centering\arraybackslash}m{3cm}}

\usepackage{blindtext}
\usepackage{geometry}
\usepackage{natbib}
\usepackage{subfig}
\usepackage{booktabs}
\usepackage{appendix}

\journal{Journal of \LaTeX\ Templates}

\newcommand{\fpp}[1]{$#1$-PPT}

\newcommand{\fp}[1]{$#1$-PT}

\SetKwBlock{addEdgesFromPQTree}{procedure addEdgesFromPQTree($G$, $D$,  $T$)}{}
\SetKwBlock{Kpt}{function partitioned-\fp{k}($G$, $k$,  $\mathcal{V}$)}{}
\SetKwBlock{Kppt}{function partitioned-\fpp{k}($G$, $k$, $\mathcal{V}$)}{}
\SetKwBlock{KpptF}{function Partitioned-\fpp{k}($G$, $\mathcal{V}$)}{}
\SetKwBlock{KpptFC}{function Partitioned-\fpp{k}($G$, $\mathcal{V}$)}{}
\SetKwFunction{AddEdgesFromPQTree}{addEdgesFromPQTree}
\SetKw{Each}{each}
\SetKw{TRUE}{TRUE}
\SetKw{FALSE}{FALSE}
\SetKw{Break}{break}
\SetKw{Not}{not}

\biboptions{authoryear}

\begin{document}

\begin{frontmatter}

\title{Empirical Evaluation of Project Scheduling Algorithms for Maximization of the Net Present Value} 

\author[1]{Isac M. Lacerda}
\ead{isac.mendes@gmail.com}

\author[1]{Eber~A.~Schmitz}
\ead{eber@nce.ufrj.br}

\author[1,2]{Jayme~L.~Szwarcfiter}
\ead{jayme@nce.ufrj.br}

\author[3]{Rosiane de Freitas}
\ead{rosiane@icomp.ufam.edu.br}

\address[1]{Federal University of Rio de Janeiro, Brazil}
\address[2]{State University of Rio de Janeiro, Brazil}
\address[3]{Federal University of Amazonas, Brazil}

\begin{abstract}
\textcolor{black}{
This paper presents an empirical performance analysis of three project scheduling algorithms dealing with maximizing projects' net present value with unrestricted resources. The selected algorithms, being the most recently cited in the literature, are:  \emph{Recursive Search} (RS), \emph{Steepest Ascent Approach} (SAA) and \emph{Hybrid Search} (HS). The main motivation for this research is the lack of knowledge about the computational complexities of the RS, SAA, and HS algorithms, since all studies to date show some gaps in the analysis. Furthermore, the empirical analysis performed to date does not consider the fact that one algorithm (HS) uses a dual search strategy, which markedly improved the algorithm’s performance, while the others don’t. In order to obtain a fair performance comparison, we implemented the dual search strategy into the other two algorithms (RS and SAA), and the new algorithms were called \emph{Recursive Search Forward-Backward} (RSFB) and \emph{Steepest Ascent Approach Forward-Backward} (SAAFB). The algorithms RSFB, SAAFB, and HS were submitted to a factorial experiment with three different project network sampling characteristics. The results were analyzed using the Generalized Linear Models (GLM) statistical modeling technique that showed: a) the general computational costs of RSFB, SAAFB, and HS; b) the costs of restarting the search in the spanning tree as part of the total cost of the algorithms; c) and statistically significant differences between the distributions of the algorithms' results.
}
\end{abstract}


\end{frontmatter}

\section{Introduction}\label{chpt:introduction}

\textcolor{black}{
Net Present Value (NPV), probably the most used method for the financial evaluation of projects, consists in calculating the sum of all discounted cash flows generated by the project activities. The premise behind the method is that the higher its NPV, the more financially attractive this project is. However, the application of this method may involve taking into account several project implementation constraints, such as: a) precedence between activities; b) dates imposed for the start or end of activities; c) $lag$ between activities; e) amount and type of resources mobilized; f) and resource renewal capacity.\\
}

\textcolor{black}{
Early discussions on how to maximize project NPV, also known as \emph{max-npv} problems,  originate from the pioneering works of \cite{B:64} and \cite{R:70}. Such contributions paved the way for the research on different classes \emph{max-npv} of problems, as shown in several works published later. The literature review of \cite{H:D:VD97} registered thirty-four works related to the subject and grouped them into six categories.\\
}

\textcolor{black}{
The category referred to as \emph{Deterministic Unconstrained max-npv} assumes (a) deterministic activity durations, (b) \emph{finish-start} precedence relations with zero time-lag, (c) deterministic cash flows and (d) deterministic discount rate. This type of \emph{max-npv} problem consists of the calculation of the set of start times of all activities that maximize a project's NPV. This will be the category discussed in this paper. The literature review of on the \emph{Deterministic Unconstrained max-npv} problem by \cite{H:D:VD97} presented seven works cataloged. However, two more recent known algorithms (useful in this category) were published after their work. These algorithms are \emph{Steepest Ascent Approach} (SAA) from \cite{SZ:01}, and \emph{Hybrid Search} (HS) from \cite{V:06}.\\
}

\textcolor{black}{
The main motivation for this research is the lack of knowledge about the computational complexities of the RS, SAA, and HS algorithms since they all show some gaps in the analysis. An additional motivation was that the empirical analysis performed to date does not consider that one algorithm (HS) uses a dual search strategy, which markedly improved the algorithm’s performance, while the others don’t. Therefore, in order to be able to obtain a fair performance comparison, we implemented the dual search strategy into the other two algorithms (RS and SAA), and the new algorithms were called \emph{Recursive Search Forward-Backward} (RSFB) and \emph{Steepest Ascent Approach Forward-Backward} (SAAFB).\\
}

\textcolor{black}{
Thus, this paper shows the results of an empirical study of the performance of the algorithms RSFB, SAAFB, and HS. They were submitted to a factorial experiment with three different project network sampling characteristics using two different performance metrics. The first metric that deals with the total cost is a proxy to the big-O complexity measure, while the second metric deals with the number of tree searches used by the algorithms. The remaining of this paper contains the following sections: \textit{The max-npv scheduling pro\-blem}; \textit{Related works}; \textit{Algorithms chosen for evaluation}; \textit{The experiment}; \textit{Results}; \emph{Analysis and discussion}; and \textit{Conclusions}.
}\unskip
\section{The max-npv scheduling problem}\label{chpt:problem}

\textcolor{black}{
The \emph{max-npv} class problem presented in this paper concerns the maximization of the Net Present Value (NPV) of projects, with precedence constraints between activities and unrestricted resources. In this sense, the information about the variables and restrictions of the problem can be displayed as a graph $G(V, E)$ where $V$ is the set of vertices and $E$ is the set of edges. Two special vertices are considered and are called initial \textit{dummy} and final \textit{dummy}, respectively demarcating the beginning and end of the project. Each vertex (representing an activity) is associated with the following attribute set: a) undiscounted cash flow (in the case of \textit{dummies} with a value of zero); b) a duration; c) a start date; d) and an end date. The vertices are arranged in ascending order, starting at the initial \textit{dummy} and ending at the final \textit{dummy}. Edges indicate a precedence relation between activities of the type \emph{finish-start} with zero $lag$. Also, the end date of the last $dummy$ must be less than or equal to the project's $deadline$.\\
}

\textcolor{black}{
Using of the three-field notation, originally presented by \cite{GLLR:79}, expanded by \cite{B:L:R83}, and specialized for the context of project scheduling by \cite {D:H97}, the problem of interest can be expressed as: $^{\circ} \mid cpm, \delta_{n}, c_{j} \mid max$-$npv$. With this notation, $^{\circ}$ indicates unrestricted resources, $cpm$ indicates precedence between \emph{finish-start} activities with zero $lag$; $\delta_{n}$ indicates that there is a $deadline$ for the project; $c_{j}$ denotes a cash flow for each activity; and $max$-$npv$ means that the objective function is to maximize the net present value.\\
}

The objective function can be given as:
\begin{equation}
    max \sum_{i=2}^{n-1} c_{i} . e^{-\alpha(s_i + d_i)}    
\end{equation}

\textcolor{black}{
in which, $c_{i}$ indicates the cash flow of each activity $i$ (at its finish time), $s_{i}$ means the start of each activity, $d_{i}$ indicates the duration of each activity and $e^{-\alpha} = 1/(1 + r)$ means the discount factor, where $r$ refers to the discount rate.\\
Subject to the following restrictions:
}

\begin{equation}
s_{i} + d_{i} \leq s_{j}; \ \forall (i, j) \in E
\end{equation}\begin{equation}
s_{1} = 0\\     
\end{equation}\begin{equation}
s_{n} \leq \delta_{n}\\
\end{equation}\begin{equation}
s_{i} \in \mathbb{N} ; i = {2,3,4...,n}\\     
\end{equation}

\textcolor{black}{
Expression (1) indicates that the objective function is to maximize the NPV. The expression (2) limits the end of any activity to be less than or equal to the start of any of its successors ($s_{j}$). The expression (3) indicates that the start time of the initial $dummy$ must be equal to zero. The expression (4) indicates that the start of the final $dummy$ ($s_{n}$) is less or equal than to the project's $deadline$ date ($\delta_{n}$). The last expression (5) indicates that starting dates must belong to the set of natural numbers.\\
}

\textcolor{black}{
It is worth mentioning that some authors have already shown that, for this class of problem, it is possible to obtain a solution in polynomial time (\citealp{CDSSWW:01,CE:85, GLP:82, JUANG:94, MPL:90, RMHTG:91, SBJ:99}). Furthermore, \cite{G:72} demonstrates that it is possible to transform the nonlinear programming problem into a linear programming problem using an event-driven approach.
}\unskip
\section{Related works}\label{chpt:related_works}

\textcolor{black}{
The literature review on the \emph{max-npv} problem by \cite{H:D:VD97} describes some of the works related to this article. This review contains seven works cataloged in the category \emph{Deterministic Unconstrained max-npv}, which is directly related to the scheduling problem of interest. However, three new works directly related to our problem were published after that. The main characteristics of these ten works are described in the sequence.\\
}

\textcolor{black}{
The first of these works is that of \cite{R:70}, one of the pioneers in promoting and formulating the maximization of net present value in projects. The author showed that, although the problem involves maximizing a nonlinear objective function (subject to linear constraints), it is possible to obtain a solution with successive applications of linear programming. The author proposed an algorithm composed of several steps but did not present any discussion either about its complexity or an empirical evaluation.\\
}

\textcolor{black}{
The second work, \cite{G:72}, demonstrates that it is possible to have a solution for the nonlinear objective function (with linear constraints) using linear programming. The author also shows that the best solution to the problem can be obtained in a (viable) tree extracted from the project network and that the search can be restricted to tree structures. The author proposes two algorithms: the first solves the problem through a single $deadline$, and the second solves the problem through several possible $deadlines$. These algorithms are related to the so-called special procedure of Markowitz for the weighted distribution problem using triangular systems of equations.\\
}

\textcolor{black}{
The third work, published by \cite{SD:86}, proposes a forecasting model for \emph{max-npv} with unrestricted resources, considering a single cash flow at project completion, characteristic of projects under lump-sum contracts. In addition, summary measures have been proposed to predict the value of \emph{max-npv}. However, the article did not propose algorithms or computational experiments.\\
}

\textcolor{black}{
In the fourth work, \cite{EH:90} demonstrated that the approaches of the papers by \cite{R:70} and \cite{G:72} can produce inconclusive results, as they do not explicitly specify start and end restrictions for scheduling. Under these arguments, \cite{EH:90} proposed an algorithm that builds tree structures iteratively and with the determination of displacement intervals. However, the paper does not present any discussion about complexity neither about computational experiments.\\
}

\textcolor{black}{
The fifth work by \cite{HG:93} presents a simplified version of the algorithm of \cite{EH:90}. The authors submitted this algorithm to a computational experiment that allowed empirical comparisons with the solutions obtained via linear programming using the software \emph{Super LINDO}. The paper does not present any discussion about complexity.\\
}

\textcolor{black}{
In the sixth work, \cite{KS:96} argued that in real projects, it is more common for cash flows to be associated with regular periods, such as months, rather than events such as the completion of activities. Thus, they proposed costs of activities divided by their duration and the formulation of the problem through integer programming. They also presented rules of a random generator of networks with results of an experiment through the software \emph{LINDO}. However, there were no discussions about complexity.\\
}

\textcolor{black}{
In the seventh work, \cite{DHV:96} describe an algorithm called \emph{Recursive Search} (RS) composed of three steps, which performs recursive searches in tree structures, assuming positive cash flow is anticipated as much as possible, and the negative ones are delayed as much as possible. This work presented a computational experiment but did not discuss any aspects of complexity.\\
}

\textcolor{black}{
In the eighth work, although the main focus was resource-constrained \emph{max-npv}, \cite{VDH:99} also proposed a refinement for the RS algorithm by adding an extra edge in the spanning tree between the last and first vertices. This extra edge favored the recursion process of the algorithm proposed initially in \cite{DHV:96}. The work of \cite{VDH:99} also presented a computational experiment but did not discuss the algorithm's complexity. However, on the version of RS with the extra edge, \cite{DH:02} and \cite{VDH:00} stated that the first two steps could be implemented with cost $O(n^2)$, but the third has unknown complexity. \\
}

\textcolor{black}{
In the ninth work, \cite{SZ:01} presented an algorithm called \emph{Steepest Ascent Approach} (SAA) with a generalized approach to the precedence of activities, admitting minimum and maximum intervals. The paper presents a computational experiment and highlights that two of its three component algorithms can be implemented respectively with cost $O(n)$ and $O(m \ log \ m)$. However, the complexity of the third component algorithm was considered an open question.\\
}

\textcolor{black}{
In the tenth and last work, \cite{V:06} proposed an algorithm called \emph{Hybrid Search} (HS) that combines RS and SAA strategies, as well as includes the ability to reverse the search and displacement direction when the design has more than half of the activities with negative cash flow. In this work, the author presented the results of a computational experiment but did not discuss any aspects of complexity.\\}

\textcolor{black}{
Table \ref{tab:related_works} presents a summary of all relevant related works, where \textbf{Algorithm} means that the work describes an algorithm, \textbf{Complexity} that it presents any complexity analysis, and \textbf{Experiment} if the paper details any form of empirical analysis.
}

\begin{table}[H]
\begin{center}
\caption{\label{tab:related_works} Summary - Related Works.}
\begin{tabular}{ l|c|c|c }
\hline
 \textbf{Authors}& \textbf{Algorithm}&\textbf{Complexity}&\textbf{Experiment}\\
 \hline
 1) \cite{R:70}   &   yes        & no            & yes       \\
 2) \cite{G:72}   &   yes        & no            & no       \\
 3) \cite{SD:86}  &   no         & no            & no       \\
 4) \cite{EH:90}  &   yes        & no            & no       \\
 5) \cite{HG:93}  &   yes        & no            & yes       \\
 6) \cite{KS:96}  &   yes        & no            & yes       \\
 7) \cite{DHV:96} &   yes        & no            & yes       \\
 8) \cite{VDH:99} &   yes        & partial       & yes       \\
 9) \cite{SZ:01}  &   yes        & partial       & yes       \\
 10) \cite{V:06}  &   yes        & no            & yes       \\

\hline

\end{tabular}
\end{center}
\end{table}
\section{Algorithms chosen for evaluation}\label{chpt:algorithms}
\textcolor{blue}{
}
\textcolor{black}{
The algorithms selected and implemented for the empirical evaluation were \emph{Recursive Search Forward-Backward} (RSFB), \emph{Steepest Ascent Approach Forward-Backward} (SAAFB), and \emph{Hybrid Search} (HS). As relates to these algorithms, it is important to point out that RSFB and SAAFB are proposed variations of the \emph{Recursive Search} (RS) and \emph{Steepest Ascent Approach} (SAA) algorithms. The following sections present the main features of each.
}

\subsection{Recursive Search Forward-Backward}
\textcolor{black}{
\emph{Recursive Search} (RS) is one of the fundamental algorithms in this work. It was originally proposed by \cite{DHV:96} and refined by \cite{VDH:99}. Its composition includes three steps. In the first step, RS creates a tree called \emph{Early Tree} (ET), with anticipation of the dates of the vertices (activities) as much as possible. Moreover, RS assumes the existence of an extra edge between the first and last vertices ($dummies$) in the ET, useful for the recursion process. In the second step, RS makes a copy of ET called \emph{Current Tree} (CT). According to the precedence constraints, the second step also delays as much as possible the vertices of CT without successors with negative cash flow. Finally, in the third step, RS recursively searches for subtrees with negative cash flow in the CT, and each of them is shifted according to the constraints.\\
}

\textcolor{black}{
Because the first two steps are meant for data preprocessing, all the interesting RS algorithmic work is done in the third step. Such a step comprises of two-component algorithms called \emph{Step\_3} and \emph{Recursion}. The first component algorithm \emph{Step\_3} (with pseudocode in Algorithm \ref{algo:Step_3_f}) is used to initiate a recursive depth-first search in the spanning tree. For this, $Step\_3$ invokes the second component algorithm \emph{Recursion} (with pseudocode in Algorithm \ref{algo:RecursionRS_f}). Thus, \emph{Recursion} can identify candidate subtrees for displacement. Each subtree identified by \emph{Recursion} is immediately displaced, and the search is restarted on the spanning tree with a new invocation to \emph{Step\_3}.\\
}

\begin{algorithm}[H]
    \label{algo:Step_3_f}
    \SetAlgoLined
    \footnotesize
     \textbf{procedure} \textit{Step\_3}() 
         $\left\{\textit{CA is a global structure}\right\}$\\
    \Indp
         \textcolor{blue}{$total\_Step\_3  = total\_Step\_3$} + 1\\
         $CA \leftarrow \varnothing$ \\  
         \underline{$SA', DC' \leftarrow$ \emph{Recursion}(1)} \\
        Report the optimal solution $DC'$\\
    \Indm

    \caption{\emph{Step\_3 - Forward.}}
\end{algorithm}

\vspace{0.3cm}

\textcolor{black}{
With respect to the complexity of RS, the authors state that the first two steps are $O(n^2)$ and the third step has unknown complexity \citep{VDH:00}. According to the authors, one point that makes this complexity an open question refers to the unknown number of times that searches are restarted in the spanning tree with invocations to the component \emph{Step\_3}. As a consequence, the overall complexity of RS is an open question.\\
}

\textcolor{black}{
Although RS is a fundamental algorithm, the version implemented and used in the experiment refers to a variation called \textit{Recursive Search Forward-Backward} (RSFB). This variation incorporates the same inversion strategy for search and displacement proposed by \cite{V:06} in the \emph{Hybrid Search} (HS) algorithm. Thus, RSFB reverses the search and shifts reference when the project has more than half of the activities with negative cash flow. In that case, RSFB creates a \emph{Late Tree} delaying the activities as much as possible and takes the $deadline$ as a reference. Then, with the inverted search, the start occurs by the final $dummy$, and the subtrees identified with positive cash flow are shifted toward the initial $dummy$.\\
}

\textcolor{black}{
In this section, the pseudocodes for the RSFB components (under Algorithms \ref{algo:Step_3_f} and \ref{algo:RecursionRS_f}) refer only to the $forward$ approach, i.e., when the project has up to half of the activities with negative cash flow. However, the pseudocodes corresponding to the $backward$ approach follow as part of  \ref{chpt:appA}. In both approaches ($forward$ and $backward$), the algorithms have their differences in underlined parts of their lines. As relates to the main terms of $Step\_3$ (Algorithm \ref{algo:Step_3_f}) the following stand out: a) $CA$ that refers to the \emph{Considered Activities} in the last search; b) $SA'$ that refers to the \emph{Set of Activities} to shift; c) and $DC'$ that refers to the \emph{Discounted Cash} flow. In addition, the variable $total\_Step\_3$ was used to count the times the search is restarted in the spanning tree. This count refers to one of the metrics discussed in the experiment section.\\
}

\textcolor{black}{
As relates to the main terms of $Recursion$ (Algorithm \ref{algo:RecursionRS_f}) the following stand out: a) the function $Compute \ v_{k*l*}$ that identifies the smallest distance between a vertex $\in SA$ and a vertex $\notin SA$; b) $f_l$, $d_k$ and $f_k$ which respectively indicate the end of vertex $l$, the duration of vertex $k$ and the end of vertex $k$; c) and $G$ which indicates the graph with all project constraints. In addition, the variable $total\_Recursion$ (line 2 in Algorithm \ref{algo:RecursionRS_f}) is used to count the number of recursive calls to the algorithm component $Recursion$. This global variable starts with a zero value only once immediately before the first call to the $Step\_3$ component. This count also refers to one of the metrics discussed in the experiment section.\\
}

\begin{algorithm}[H]
    \label{algo:RecursionRS_f}
     \footnotesize
     \textbf{function} \emph{Recursion}($newnode$)  \quad $\left\{\textit{CA, CT are global structures}\right\}$ \ \quad \quad \quad \quad  
     \\
    \SetAlgoNoLine
     \Indp
         \textcolor{blue}{$total\_Recursion  = total\_Recursion$} + 1\\
         $SA \leftarrow \left\{newnode\right\}; DC \leftarrow DC_{newnode}; CA \leftarrow CA + newnode$ \\ 
    
         \For{\Each$(i | i \notin CA \ \textbf{\emph{and}} \ i \ \underline{\emph{succeeds} \ newnode \in CT})$}
         {
            $SA', DC' \leftarrow Recursion(i)$   \\
            \eIf{$DC' \geq 0$}{$SA  \leftarrow SA + SA';DC  \leftarrow DC + DC'$}
            {
                $\underline{CT \leftarrow CT - (newnode, i)}$ \\
                $\emph{Compute} v_{k*l*} = min \left\{\underline{f_l - d_k - f_k} \right\};$ $\underline{CT \leftarrow CT + (k*,l*)}$ \\
                $\quad \quad \quad \quad \quad \quad \quad \quad ^{\underline{(k*,l*) \in G}}$\\
                $\quad \quad \quad \quad \quad \quad \quad \quad ^{\underline{k* \in SA}}$\\
                $\quad \quad \quad \quad \quad \quad \quad \quad ^{\underline{l* \notin SA}}$\\ 
                $\forall \ j \in SA' : \underline{f_j \leftarrow f_j + v_{k*l*}}$ \\
                \emph{Step\_3}()
            }
        }
        \For{\Each$(i | i \notin CA \ \textbf{\emph{and}} \ i \ \underline{\emph{precedes} \ newnode \in CT})$}
        {
             $SA', DC' \leftarrow \emph{Recursion}(i)$\\   
             $SA \leftarrow SA + SA';DC \leftarrow DC + DC'$
        }
    \Return(\emph{SA, DC})\\
    \Indm
    \caption{\emph{Recursion - Forward.}}
\end{algorithm}

\subsection{Steepest Ascent Approach Forward-Backward}
\textcolor{black}{
Proposed by \cite{SZ:01}, \emph{Steepest Ascent Approach} (SAA) is the second fundamental algorithm in this work. Although his approach generalizes the precedence relationship between activities, an adaptation is easily performed, considering only the type \emph{finish-start} and zero $lag$, as has been done in similar work \citep{VDH:00}. Its organization includes the three-component algorithms \emph{Steepest\- Ascent\- Direction\-} (SAD), \emph{Vertex\- Ascent\-} (VA), and \emph{Steepest\- Ascent\- Procedure\-} (SAP). Thus, the SAD component algorithm iteratively searches for subtrees with negative cash flow in the spanning tree. The VA component algorithm identifies destinations\- for the subtrees found with SAD, considering the closest constraints. The third SAP component algorithm synthesizes the presented strategy, including calls to the SAD and VA components. With this, SAP calls SAD, and when subtrees are identified, VA is executed to perform the displacements. This way, SAP iterates as long as subtrees are identified with SAD.\\
}

\textcolor{black}{
With respect to complexity, the authors claim that SAD can be implemented in $O(n)$ and VA in $O(m \ log \ m)$ which is equivalent to $O(m \ log \ n^2)$ = $O( 2 \ m \ log \ n)$ = $O(m \ log \ n)$. However, the lack of knowledge of the number of times that searches in the spanning tree are restarted is an open question, as it also occurs in RS. Hence, the overall complexity of SAA remains an open question.\\
}

\textcolor{black}{
Although SAA is another fundamental algorithm, the version implemented and used in the experiment refers to a variation called \emph{Steepest Ascent Approach Forward-Backward} (SAAFB). Similar to what was proposed in RSFB, SAAFB also incorporates the inversion strategy for searching and shifting of the HS \citep{VDH:00}. The pseudocodes of the SAAFB components (Algorithms \ref{algo:SAD_f}, \ref{algo:VA_f} and \ref{algo:SAP_f}) are presented in the $forward$ approach. However, versions in the $backward$ follow as part of \ref{chpt:appA}. The differences between these approaches are highlighted in underlined passages in the pseudocodes.
}
\SetKwComment{Comment}{/* }{ */}
\begin{algorithm}
    \footnotesize
	 $\textbf{function} \ \emph{SAD}()$  \quad $\left\{ST(V_{st}, E_{st})\textit{ is a global structure}\right\}$\\
     \SetAlgoLined
     \Indp
        \SetAlgoNoLine
         $Z \leftarrow \varnothing$; $V \leftarrow V_{st}$\\ 
         $\forall \ i \in V \ \mathbf{do} \ C(i) \leftarrow  {i}$; $\phi_{i} \leftarrow -\alpha \ c_{i} \ e^{-\alpha^{(S_{i} + d_{i})}}$\\
        \While{V $\neq $ $\left\{ 1 \right\}$}{
                \eIf{$(V \emph{ has a \underline{node source} $i$ $\neq 1$}) \And (\emph{at most \underline{one successor}}\ j)$}{
                \textcolor{blue}{$iteration\_SAD  = iteration\_SAD$} + 1\\
                $C(j) \leftarrow  C(j) + C(i)$;
                 $\phi_{j} \leftarrow \phi_{j} + \phi_{i}$;
                 $V \leftarrow V - i$ \\
                }{
                 \If{$(V \emph{ has a \underline{node sink} $j$ $\neq 1$}) \And (\emph{only \underline{one predeccessor}}\ i)$}{
                     \textcolor{blue}{$iteration\_SAD  = iteration\_SAD$} + 1\\
                     $\mathbf{if} \ \underline{\phi_{j} > 0} \ \mathbf{then} \ Z \leftarrow Z + C(j)$ \\
                    $\mathbf{else}$ \ $\phi_{i} \leftarrow \phi_{i} + \phi_{j}$; \ $C(i) \leftarrow C(i) + C(j)$; \ 
                    $V \leftarrow V - j$\\
                    }
                }
           }
        $\Return(Z)$\\
    \Indp
     \caption{\emph{Steepest Ascent Direction} (SAD) - \emph{Forward.}}
     \label{algo:SAD_f}
\end{algorithm}

\textcolor{black}{
The main terms of the SAD component are: a) $ST$ as the spanning tree, with $V_{st}$ being the vertices and $E_{st}$ the edges; b) $Z$ is the set of candidate vertices for displacement; c) $C(i)$ a vector used to group vertices without successors or predecessors; d) and $\phi_{i}$ refers to the partial derivative of the discounted cash flow of each activity $i$.\\
}

\begin{algorithm}[H]
    \label{algo:VA_f}
     \SetAlgoNoLine
     \footnotesize
     \textbf{function} \emph{VA}(\emph{S, Z}) \quad $\left\{\textit{ST, G are global structures}\right\}$\\
     \Indp 
         $\forall \ (i,j) \in ST \ | \ (j \notin C(i)) \And (i \notin C(j)) $ :  $ST \leftarrow ST - (i, j)$ \\
          \While{$Z \neq \varnothing$}{
          $\emph{Compute} v_{k*l*} = min \left\{\underline{S_l - S_k - d_k} \right\}$\\
           $\quad \quad \quad \quad \quad \quad \quad \quad \underline{^{(k*,l*) \in G}}$\\
           $\quad \quad \quad \quad \quad \quad \quad \quad \underline{^{k* \in Z}}$\\
           $\quad \quad \quad \quad \quad \quad \quad \quad \underline{^{l* \notin Z}}$\\
           
           Take the set $C(j) \in Z$ where $k*$ is contained\\
           
           $\forall \ i \in C(j) : S_i \leftarrow \underline{S_i + v_{k*l*}}$\\
            $Z \leftarrow Z - C(j);$ \
            $ST \leftarrow$ \underline{$ST + (k*,l*)$}
           }
        \Return(\emph{S})\\
    \Indm
     \caption{\emph{Vertex Ascent} (VA) - \emph{Forward.}}
\end{algorithm}

\textcolor{black}{
The main terms of the VA component (Algorithm \ref{algo:VA_f}) are: a) $S$ as the vector that contains the schedule for the start of the activities of the spanning tree; b) and $Compute \ v_{k*l*}$ as the function that calculates the shortest distance between $k \in Z$ and $l \notin Z$. The main terms contained in the SAP component (Algorithm \ref{algo:SAP_b}) have already been highlighted.\\}

\begin{algorithm}[H]
     \SetAlgoNoLine
     \footnotesize
     \textbf{procedure} \emph{SAP}() \quad $\left\{\textit{ST is a global structure}\right\}$\\
     \Indp
         $S, ST \leftarrow $ \underline{Determine the Early Schedule ($S$)} as a vector and a corresponding initial Spanning Tree ($ST$) through the original graph $G$. \\
         \textcolor{blue}{$total\_SAD  = 0$}\\
         $Z \leftarrow$ \emph{SAD}()\\
         \While{$Z \neq \varnothing$}{
           \textcolor{blue}{$total\_SAD  = total\_SAD$} + 1\\
           $S \leftarrow$ \emph{VA(S, Z)}\\
           $Z \leftarrow$ \emph{SAD}()\\
           }
           Report the optimal solution $S$\\
     \Indm  
     \caption{\emph{Steepest Ascent Procedure} (SAP) - \emph{Forward.}}
     \label{algo:SAP_f}
\end{algorithm}

\textcolor{black}{Finally, the variables $iteration\_SAD$ (Algorithm \ref{algo:SAD_f}) and $total\_SAD$ (Algorithm \ref{algo:SAP_f}) refer res\-pectively to the total number of iterations performed in the search for vertices that meet the conditionals of lines 5 and 9 (SAD), and the total number of searches initiated in the spanning tree with the invocation of SAD. Such counts are metrics discussed in the experiment section.
}

\subsection{Hybrid Search}
\textcolor{black}{
Proposed by \cite{V:06}, \emph{Hybrid Search} (HS) is the last algorithm considered in this work. Its approach combines RS and SAA strategies and is the pioneer algorithm in the inversion of search and displacement when more than half of the activities have negative cash flow. HS includes three-component algorithms called \emph{Recursion}, \emph{Shift\_activities} and \emph{Hybrid Recursive Search} (HRS). Thus, $Recursion$ uses recursive depth-first search to identify candidate subtrees for displacement (with pseudocode in Algorithm \ref{algo:RecursionHS_f}). The \emph{Shift\_activities} component (with pseudocode in Algorithm \ref{algo:Shift_activities_f}) finds destinations and performs the displacements of the subtrees identified in the last depth-first search. The \emph{HRS} component (with pseudocode in Algorithm \ref{algo:HRS_f}) summarizes the entire scheduling approach, invoking the \emph{Recursion} and \emph{Shift\_activities} components.
}
\begin{algorithm}[H]
     \SetAlgoNoLine
     \footnotesize
     \textbf{function} \emph{Recursion}($newnode$) \quad $\left\{\textit{CA, ST, SS are global structures}\right\}$\\
     \Indp
         \textcolor{blue}{$total\_Recursion  = total\_Recursion$} + 1\\
         $SA \leftarrow \left\{newnode\right\}; DC \leftarrow DC_{newnode}; \ CA \leftarrow CA + newnode$\\
         \For{\Each$(i | i \notin CA \ \textbf{\emph{and}} \ \underline{i \ \emph{succeeds} \ newnode \in ST})$}{
            $SA', DC' \leftarrow \emph{Recursion}(i)$\\
            \eIf{\underline{$DC' \geq 0$}}{$SA \leftarrow SA + SA';DC \leftarrow DC + DC'$}
            {
                $ST \leftarrow \underline{ST - (newnode, i)}; \ SS \leftarrow SS + SA'$
            }
         }
        \For{\Each$(i | i \notin CA \textbf{\emph{and}} \underline{i \emph{precedes} newnode \in ST)}$}
         {
             $SA', DC' \leftarrow \emph{Recursion}(i)$\\ 
             $SA \leftarrow SA + SA';DC \leftarrow DC + DC'$\\
         }
         \Return(\emph{SA, DC})\\
    \Indm
    \caption{\emph{Recursion} de HS - $Forward$.}
    \label{algo:RecursionHS_f}
\end{algorithm}

\textcolor{black}{
Although the subtree search strategy is also recursive like RS, the HS algorithm can identify several subtrees before performing displacements, just like the SAA algorithm (but the latter in an iterative way). Therefore, several subtrees may have been identified in the last search performed when HS starts to perform displacements, similar to SAA.
\\}

\begin{algorithm}[H]
     \SetAlgoNoLine
     \footnotesize
     \textbf{procedure} \emph{Shift\_activities}() \quad
     $\left\{\textit{SS, ST, and G are global structures}\right\}$\\
     \Indp
         $Z \leftarrow \varnothing; \ \forall \ i \in SA \ | \ SA \in SS: Z \leftarrow Z + i$ \\
         
         \While{$Z \neq \varnothing$}{
    
           $\emph{Compute} v_{k*l*} = min \left\{\underline{s_l - s_k - d_k} \right\}$\\
           $\quad \quad \quad \quad \quad \quad \quad \quad \underline{^{(k*,l*) \in G};}$\\
           $\quad \quad \quad \quad \quad \quad \quad \quad \underline{^{k* \in Z;}}$\\
           $\quad \quad \quad \quad \quad \quad \quad \quad \underline{^{l* \notin Z}}$\\
           
            $\forall \ i \in SA | k* \in SA : s_i \leftarrow \underline{s_i + v_{k*l*}}$ \textbf{and} $Z \leftarrow Z - i$\\
            $ST \leftarrow \underline{ST + (k*,l*)}$\\
           }
     \Indm
     \caption{\emph{Shift\_activities} - \emph{Forward.}}
     \label{algo:Shift_activities_f}
\end{algorithm}

\textcolor{black}{
There are no complexity considerations for HS, making this aspect an open question. The main terms of the $Recursion$ are: a) $SA$ as the candidate set of activities; b) $DC$ as discounted cash flow; c) $CA$ as activities considered in the search; d) $ST$ as spanning tree; e) and $SS$ as a set of set of activities. The main terms of the $Shift\_activities$ component are: a) $Z$  as the vertices that must be shifted; b) and the $Compute \ v_{k*l*}$ as a function to calculate the shortest distance.\\
}

\begin{algorithm}[H]
     \SetAlgoLined
     \footnotesize
     \textbf{procedure} \emph{HRS}() \quad
     $\left\{\textit{CA and SS are global structures}\right\}$\\
     \Indp
         \textcolor{blue}{$total\_HRS  = total\_HRS$} + 1\\
         $CA \leftarrow SS \leftarrow \varnothing$\\
         $SA, DC' \leftarrow$ \underline{\emph{Recursion}(1)}\\
         {\textbf{if} $SS \neq \varnothing$ \textbf{then}\\}
         \quad  {\emph{Shift\_activities}()\\ 
         \quad   \emph{HRS}()\\} {\textbf{else} \ Report the optimal solution $DC'$}\\
    \Indm
    \caption{\emph{Hybrid Recursive Search} (HRS) - \emph{Forward.}}
    \label{algo:HRS_f}
\end{algorithm}

\vspace{0.3cm}

\textcolor{black}{
Finally, the variables $total\_Recursion$ (Algorithm \ref{algo:RecursionHS_f}) and $total\_HRS$ (Algorithm \ref{algo:HRS_f}) refer respectively to the total number of recursive calls in searches by displacement candidate subtrees and the total number of searches initiated in the spanning tree. These counts are metrics discussed in the experiment section.
}

\subsection{$Compute v_{k*l*}$}
\textcolor{black}{
\emph{Compute} $v_{k*l*}$ refers to a function contained in the fundamental algorithms and in all the algorithms (variations) implemented for the experiment of this research. Its purpose is to identify the shortest distance between a vertex that must be moved to a vertex that must not. Although this purpose is at the heart of the logic for scheduling, the fundamental algorithms (RS, SAA, and HS) treat \emph{Compute} $v_{k*l*}$ as a black box. In other words, the original works referring to fundamental algorithms did not present open pseudocode for this function. For this reason, the $forward$ version of \emph{Compute} $v_{k*l*}$ implemented together with the RSFB, SAAFB and HS algorithms remains explicit in Algorithm \ref{algo:Compute_vkl_f}. The pseudocode corresponding to the $backward$ approach is contained in \ref{chpt:appA}. Underlined points highlight the differences between the $forward$ and $backward$ approaches (Algorithm \ref{algo:Compute_vkl_f}).\\
}

\begin{algorithm}[H]
     \SetAlgoNoLine
     \footnotesize
     \textbf{function} $Compute \ v_{k*l*}(Z)$\\
     \Indp
         $v_{k*l*} \leftarrow \delta$\\
         $k \ \leftarrow \varnothing$; $l \ \leftarrow \varnothing$\\
         \textbf{for} $node \in Z$ \textbf{do}\\
            \Indp
            \textbf{if} $k = \varnothing$ \textbf{then}$ \ k \leftarrow node$\\
            \textbf{for} $suc \in \ \underline{successors of}$ \ node \ \textbf{do}\\
            \Indp
                $\textcolor{blue}{\emph{edge\_checked}} \leftarrow \textcolor{blue}{\emph{edge\_checked}} + 1$\\
                \textbf{if} suc $\notin Z$ \textbf{do}\\
                    \Indp
                    
                        \eIf{$\underline{s_l - s_k < 0}$}{$current\_min = \underline{s_l - s_k - (-\delta)}$}{$current\_min = \underline{s_l - s_k - d_k}$}
                        \eIf{$current\_min < v_{k*l*}$}{
                            $v_{k*l*} \leftarrow current\_min$\\
                            \eIf{\underline{$node$\ \emph{is\ the\ last\ node}}}{
                            $\underline{l \leftarrow 1}$}
                            {$\underline{l \leftarrow suc}$} 
                             $k \leftarrow node$}{
                            \textbf{if} $l = \varnothing$ \textbf{then}$ \ l \leftarrow suc$\\ 
                             }
                        
                    \Indm
                    
            \Indm
     \Indm
     \textbf{return} ($k, l, v_{k*l*}$)\\
     
    \caption{\emph{Compute} $v_{k*l*}$ - $Forward.$}
    \label{algo:Compute_vkl_f}
\end{algorithm}

\vspace{0.3cm}

\textcolor{black}{
As all the main terms contained in $Compute \ v_{k*l*}$ have already been explained in the descriptions of the algorithms, only one remark about the variable $edge\_checked$ should be made. This variable refers to the count of edges checked in moving into the nearest constraints. It is treated as a global variable started only once (with a zero value) at the instant when any of the algorithms are also started. Thus, the section dealing with the experiment discusses the count of checked edges as one of the metrics.
}\unskip
\section{ The experiment}\label{chpt:experiment}
\subsection{Independent variables: the experimental factors}

\textcolor{black}{
The experiment aimed to investigate the absolute and relative computational performance of the RSFB, SAAFB, and HS algorithms under different characteristics of the project networks. These characteristics or experimental factors constitute independent variables of the experiment. Table \ref{tab:factors} shows the experimental factors and their respective meanings.
}

\begin{table}[H]
\begin{center}
\caption{\label{tab:factors} Experimental Factors.}
\begin{tabular}{ c|c|l}
\hline
 \textbf{Factor}    &\textbf{Code} & \textbf{Description}                                             \\
 \hline
 \emph{vertices}       & $f_1$ & Number of vertices of the network graph.                     \\
 \emph{layers}      & $f_2$ & Number of layers of the network graph.                           \\
 \emph{maxDegree}   & $f_3$ & Maximum degree (in and out) of the vertices of the network graph.\\ 
 \emph{discRate(\%)}& $f_4$ & The discount rate used in the project.                            \\   
 \emph{percNeg(\%)} & $f_5$ & Percentage of activities with negative cash flow.                 \\ 
 \emph{cpMult}      & $f_6$ & Project deadline as a multiple of the critical path duration.                  \\
 \emph{edges}      & $f_7$ & Number of edges of the network graph.                  \\
\hline
\end{tabular}
\end{center}
\end{table}

\subsection{Dependent variables: direct metrics }

\textcolor{black}{
The dependent variables of the experiment were selected in such a way as to provide a measure of the computational cost of the selected algorithms. The dependent variables were defined by two metrics: the first metric, called \emph{computational cost}, represents the total computational cost incurred by the algorithms to evaluate the optimal schedules. It is obtained by adding the number of iterations (in iterative algorithms), or the number of recursive calls (in recursive algorithms) to the number of edges checked in the search for the shortest displacement distance of subtrees. The formula of the \emph{computational cost} metric is displayed below:
}

\begin{equation}
\label{eq:computational_cost_RSFB_HS}
 \emph{computational cost}  = total\_Recursion + edge\_checked
\end{equation}
\begin{equation}
\label{eq:computational_cost_SAAFB}
 \emph{computational cost}  = iteration\_SAD + edge\_checked
 \end{equation}

\textcolor{black}{
In the RSFB and HS algorithms, the composition of \emph{computational cost} refers to the one indicated in equation \ref{eq:computational_cost_RSFB_HS}. In this case, the variable $total\_Recursion$ (line 2 of the pseudocodes in Algorithms \ref{algo:RecursionRS_f} and \ref{algo:RecursionHS_f}) counts the number of recursive calls and the variable $edge\_checked$ (line 7 of the pseudocode in Algorithm \ref{algo:Compute_vkl_f}) counts the edges checked. In the SAAFB algorithm, the composition of \emph{computational cost} refers to the one indicated in the equation \ref{eq:computational_cost_SAAFB}. In this case, the variable $iteration\_SAD$ (lines 6 and 10 of the pseudocode in Algorithm \ref{algo:SAD_f}) counts the iterations performed and the variable $edge\_checked$ (also in line 7 of the pseudocode in Algorithm \ref{algo:Compute_vkl_f}) counts the edges checked.\\
}

\textcolor{black}{
The second metric, called \emph{restarted search}, represents the number of times a new search is restarted in the spanning tree. In the case of the RSFB algorithm, \emph{restarted search} is represented by the variable $total\_Step\_3$, indicated in line 2 of the pseudocode in Algorithm \ref{algo:Step_3_f}. This variable accounts for all calls to the $Step\_3$ component of RSFB. In the case of the SAAFB algorithm, \emph{restarted search} is represented by the variable $total\_SAD$, indicated in line 3 of the pseudocode in Algorithm \ref{algo:SAP_f}. This variable accounts for all calls to the $SAD$ component of SAAFB. In the case of the HS algorithm, \emph{restarted search} is represented by the variable $total\_HRS$, indicated in line 2 of the pseudocode in Algorithm \ref{algo:HRS_f}. This variable accounts all calls to the $HRS$ component of HS.\\
}

\subsection{Dependent variables: the upper-bound metric}\label{sec:max-cost}

\textcolor{black}{
The most common computational measures of algorithm cost are a) asymptotic upper bound ($O$), b) asymptotic lower bound ($\Omega$), c) and asymptotic upper and lower bound ($\Theta$). In this sense, this experiment was concerned with finding a statistical asymptotic upper bound ($O$) in the form of an empirical maximum cost function as an approximate return of the maximum computational cost as a function of an experimental factor, i.e., $maxCost(factor= value_{factor})$. In this case, the function returns the maximum computational cost obtained in the experiment when the specific $factor$ assumes $value_{factor}$.\\
}

\textcolor{black}{
Since each one of the experimental factors $f_i$ assumes values in a discrete set $Df_i$, the whole sample space of the experiment is given by: $$S=\prod_{i=1}^{i=6}Df_i$$}

\textcolor{black}{
So, the sample space of the experiment is the relation containing all tuples that can be formed with all possible  combinations of experimental factor values, as follows: \\ $$S=\{(vf_1,vf_2,\dots,vf_n)|(vf_1 \in Df_1,vf_2\in Df_2,\dots,vf_n\in Df_n)\}$$.}

\textcolor{black}{
Thus, on the sample space, the computational cost function is $S \rightarrow R^+$ and a sample subspace is 
$S_{f_i,vf_i} =\{ s \in S | s(\dots,f_i=vf_i,\dots\}$, which is subset of $S$, containing all tuples where the factor $f_i$ assumes the value $vf_i$. In the same way, the empirical computational cost function is $Cost(vf_1,vf_2,\dots,vf_n)=Cost(f_1=vf_1,f_2=vf_2,\dots,f_n=vf_n)$, which the experimental value obtained for each element of the sample space $S$. Finally, the conditional empirical function $maxCost(f_i,vf_i)$ can be defined as: $maxCost(f_i,vf_i) = max(Cost(s_i| s_i \in S_{f_i,vf_i}))$.
}
\subsection{Characteristics of the experimental samples}

\textcolor{black}{
The experiment ran the three algorithms in three batches comprising 14,000 project network instances (5,000 in Sample 1, 5,000 in Sample 2, and 4,000 in Sample 3). The instances were created from a random network generator, developed exclusively for the experiment, inspired by \cite{ER:60}. The random instance generator obtains random instances from sampling values in the ranges of values defined for the independent variable parameters, as shown in Table \ref{tab:lotes}.\\
}

\begin{table}[H]
\begin{center}
\caption{\label{tab:lotes} Random Sampling.}
\begin{tabular}{ c|c|c|c }
\hline
 \textbf{Parameter}& \textbf{Sample 1}&\textbf{Sample 2}&\textbf{Sample 3}\\
 \hline
 \emph{vertices} &         16..80          & 16..320       & 16..320       \\
 \emph{layers}&         2..$vertices-1$        & 2..$vertices-1$      & 2             \\
 \emph{maxDegree}&      2 or 3          & 2 or 3        & $(vertices-2)/2$       \\
 \emph{discRate}(\%)&   1..20           & 1..20         & 1..20         \\
 \emph{percNeg}(\%)&    0,10,20,..100   & 0,10,20,..100 & 0,10,20...50  \\ 
 \emph{cpMult}&         1..2            & 1..2          & 1..2          \\
 \emph{cashFlow}&       -100..100       & -100..100     & -100..100      \\
 \emph{activityDur}&    5..10           & 5..10         & 5..10         \\

\hline

\end{tabular}
\end{center}
\end{table}

\textcolor{black}{
Samples 1 and 2 differ only in the number of vertices of the networks, the purpose being to evaluate the behavior of the algorithms on two sets of project networks: small/medium networks (Sample 1) and large networks (Sample 2). \\
}

\textcolor{black}{
Sample 3, on the other hand, was created to evaluate the behavior of algorithms in graphs with only two layers (disregarding \textit{dummies}), in which all vertices of the first layer are connected to all vertices of the second layer. Such a configuration generated graphs with the maximum number of edges, i.e., complete bipartite digraphs (disregarding \textit{dummies}). It provided a sample of data designed to stress the algorithms with a higher number of edges in the search subtrees to be displaced.\\
}

\textcolor{black}{
The algorithms were coded in \emph{Python 3} and run on a personal computer, with a 2.50 GHz \emph{core i5} processor, with 32 GB of RAM, running under \emph{Windows} 10. It is worth noting that although there are instances available in the literature, it was decided to create and use a specific generator for this experiment to establish ranges of parameter values with a gradual increase and flexibility.
}

\subsection{Statistical tools: GLM}

\textcolor{black}{
The classical linear regression was the first data analysis technique considered. However, since the assumptions of the classical linear regression approach, such as normality of the residuals of the distributions and homoscedasticity, were not satisfied, even with Box-Cox and logarithmic transformations, a different tool had to be used - the generalized regression. This approach is part of the group called Generalized Linear Models (GLM), introduced by \cite{N:W72}. In this case, the generalized regression approach admits linear and non-linear models, besides dependent variables with distributions such as Bernoulli, Poisson, Poisson-Gamma, and Gaussian (normal).\\
}

\textcolor{black}{
According to \cite{FP:19}, the proper definition of a generalized linear model must consider the dependent variable's characteristic. The characteristics of two types of dependent variables presented by \cite{FP:19} are worth noting. In the first one, the generalized model is called linear type when the dependent variable is quantitative and adherent to the normal distribution. In the second one, when the dependent variable is quantitative, being a count data (integer and non-negative values), not adherent to the Gaussian, it is considered Poisson or Poisson-Gamma. It is worth mentioning that the difference between a Poisson distribution and a Poisson-Gamma distribution is the long tail to the right of the second distribution (Poisson-Gamma), characterizing overdispersion of the data. In other words, when the variance is statistically greater than the mean (with count data), the Poisson-Gamma distribution should be chosen. In this case, the generalized model is called a negative binomial.\\
}

\textcolor{black}{
In this work, all results were analyzed using generalized models with negative binomial distribution (Poisson-Gamma).
}

\section{Results: preliminary data analysis}

The preliminary analysis of the experimental results took the following steps: (1) generate a summary analysis of the data, (2) check for the similarity between the empirical distribution, and (3) verify the degree of correlation between the factors and the dependent variables.

\subsection{Dependent variable summary measures}

Tables \ref{tab:summaryAm1}, \ref{tab:summaryAm2} and \ref{tab:summaryAm3} present the minimum, first quartile, median, mean, third quartile and maximum, by metric and sample type. 

\begin{table}[H]
\centering
\tiny
\caption{Summary Sample 1.} 
\label{tab:summaryAm1}
 \subfloat[\textit{computational cost} metric.]{
 \centering
    \begin{tabular}{lrrrrrr}
    \hline
     Algo & Min. & 1st Q. & Med. & Mean & 3rd Q. & Max. \\ 
     \hline
      HS & 18 & 93 & 182 & 301 & 375 & 4308 \\ 
      RSFB & 18 & 77 & 261 & 614 & 685 & 25587 \\ 
      SAAFB & 17 & 91 & 180 & 298 & 371 & 4298 \\ 
     \hline
    \end{tabular}
 }
 \hspace{0.1cm} 
 \subfloat[\textit{restarted search} metric.]{
 \centering
    \begin{tabular}{lrrrrrr}
    \hline
     Algo & Min. & 1st Q. & Med. & Mean & 3rd Q. & Max. \\ 
     \hline
     HS & 1 & 2 & 3 & 3 & 4 & 15 \\ 
     RSFB & 1 & 3 & 9 & 15 & 20 & 339 \\ 
     SAAFB & 1 & 2 & 3 & 3 & 4 & 15 \\ 
   \hline
    \end{tabular}
 }
\end{table}

\begin{table}[H]
\centering
\tiny
\caption{Summary Sample 2.} 
\label{tab:summaryAm2}
 \subfloat[\textit{computational cost} metric.]{
 \centering
    \begin{tabular}{lrrrrrr}
    \hline
     Algo & Min. & 1st Q. & Med. & Mean & 3rd Q. & Max. \\ 
     \hline
       HS & 18 & 310 & 978 & 2795 & 3008 & 58409 \\ 
       SAAFB & 17 & 307 & 973 & 2791 & 3003 & 58391 \\ 
     \hline
    \end{tabular}
 }
 \hspace{0.1cm} 
 \subfloat[\textit{restarted search} metric.]{
 \centering
    \begin{tabular}{lrrrrrr}
    \hline
     Algo & Min. & 1st Q. & Med. & Mean & 3rd Q. & Max. \\ 
     \hline
    HS & 1 & 2 & 3 & 4 & 5 & 26 \\ 
    RSFB & 1 & 2 & 3 & 4 & 5 & 26 \\   
     \hline
    \end{tabular}
 }
\end{table}

\begin{table}[H]
\centering
\tiny
\caption{Summary Sample 3.} 
\label{tab:summaryAm3}
 \subfloat[\textit{computational cost} metric.]{
 \centering
    \begin{tabular}{lrrrrrr}
    \hline
     Algo & Min. & 1stQ. & Med. & Mean & 3rdQ. & Max. \\ 
     \hline
        HS & 18 & 2626 & 31749 & 351981 & 271013 & 10186880 \\ 
        RSFB & 17 & 2620 & 31725 & 351949 & 270980 & 10186639 \\
     \hline
    \end{tabular}
 }
  \hspace{0.1cm} 
 \subfloat[\textit{restarted search} metric.]{
 \centering
    \begin{tabular}{lrrrrrr}
    \hline
     Algo & Min. & 1stQ. & Med. & Mean & 3rdQ. & Max. \\ 
     \hline
        HS & 1 & 8 & 21 & 32 & 48 & 241 \\ 
        RSFB & 1 & 8 & 21 & 32 & 48 & 241 \\ 
     \hline
    \end{tabular}
 }
\end{table}

\subsection{Empirical result distribution types}

\begin{figure}[H]
    \centering
    \includegraphics[width=10cm]{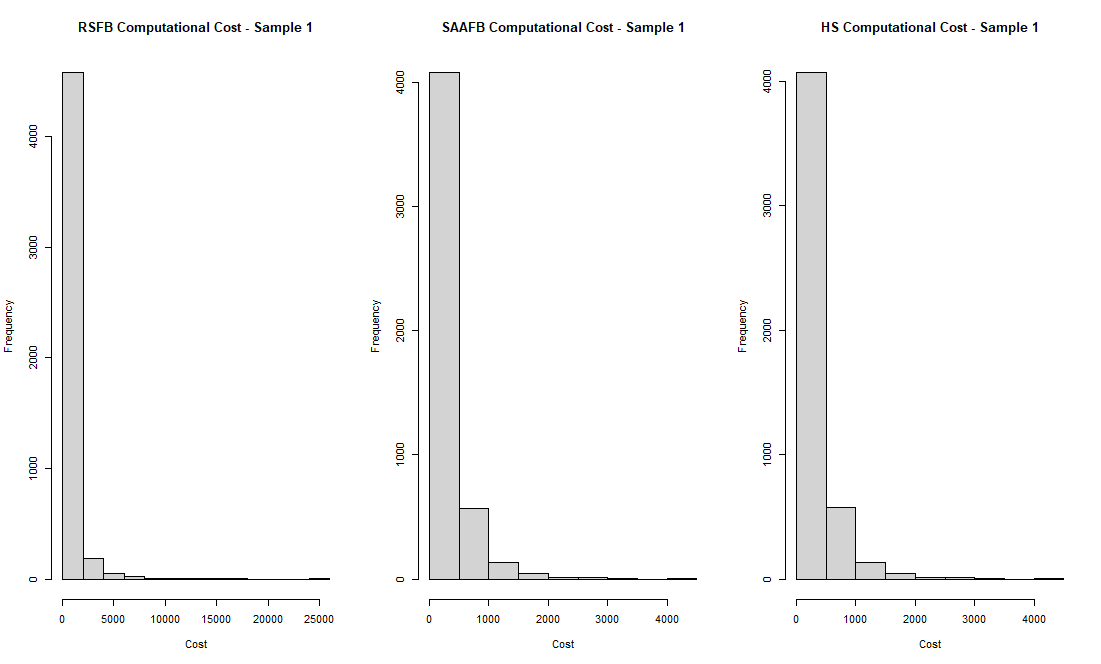}    \caption{\label{fig:histAm1} Distributions of \textit{computational cost} - Sample 1.}
\end{figure}

\begin{figure}[H]
    \centering
    \includegraphics[width=10cm]{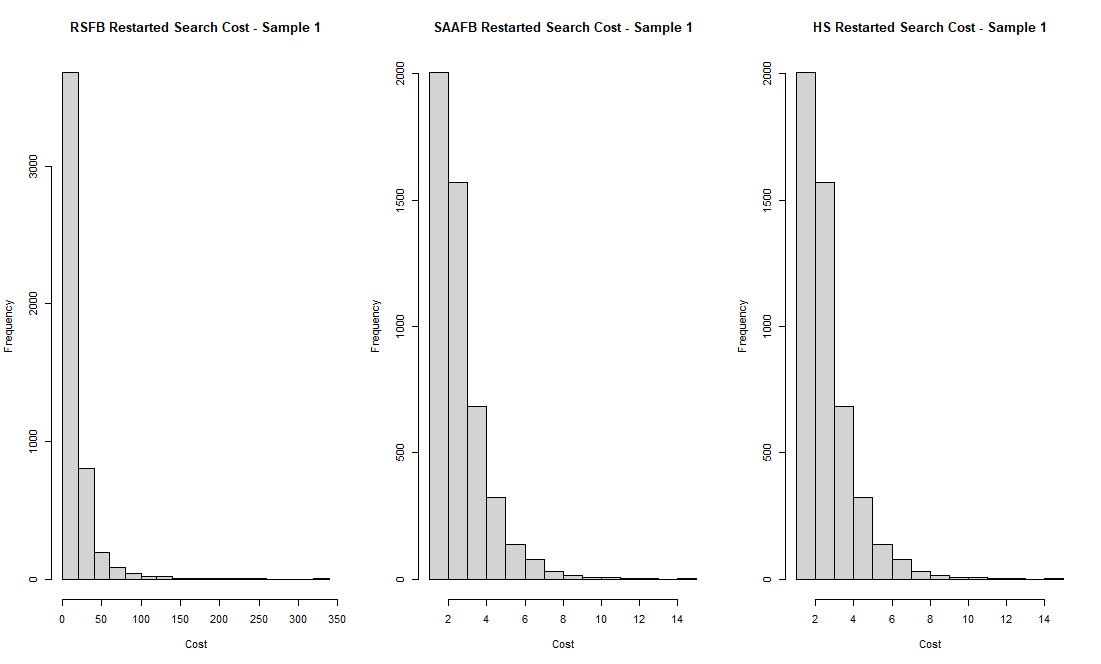}    \caption{\label{fig:histAm1_search} Distributions of \textit{restarted search} - Sample 1.}
\end{figure}

Figures \ref{fig:histAm1} and \ref{fig:histAm1_search} show that the distributions have a greater concentration of results on the left and present an exponential drop with a long tail on the right. This pattern suggests that the distributions can be Poisson or Poisson-Gamma, as pointed by \cite{FP:19}, considering that these variables are counting data.


\begin{figure}[H]
    \centering
    \includegraphics[width=11cm]{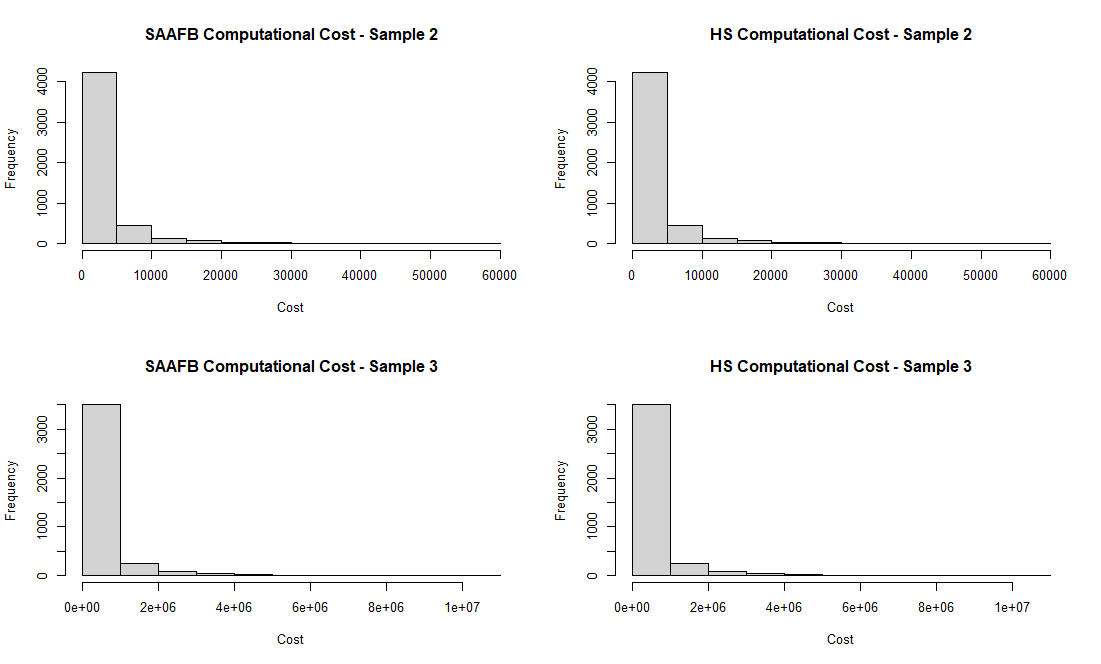}    \caption{\label{fig:histAm12} Distribution of \textit{computational cost} - Samples 2 and 3.}
\end{figure}

\begin{figure}[H]
    \centering
    \includegraphics[width=11cm]{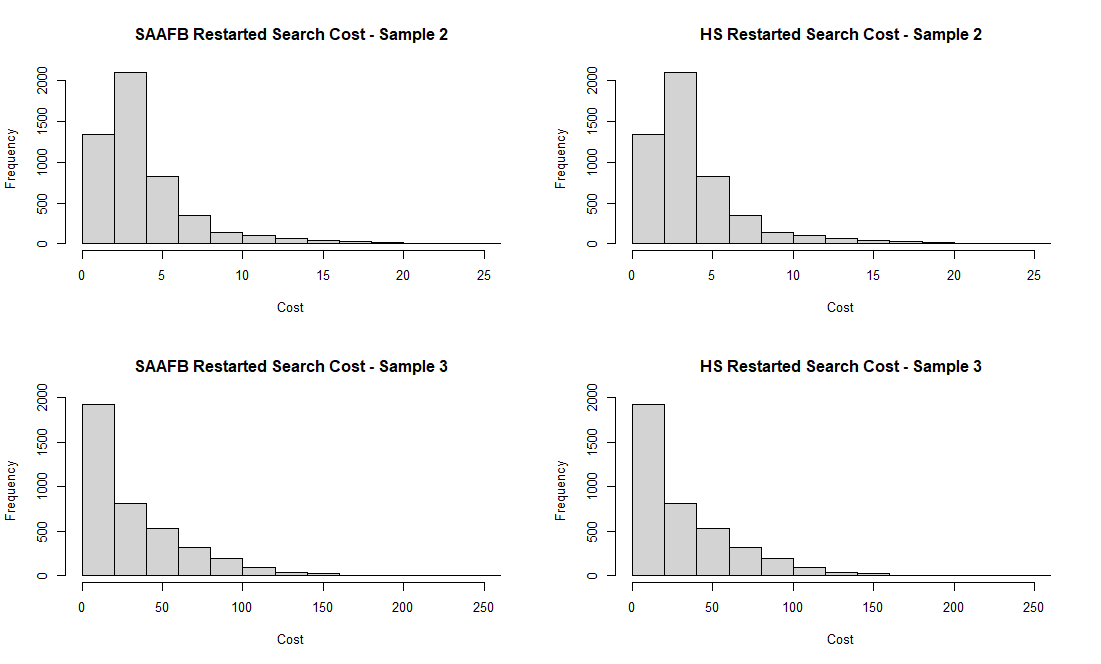}    \caption{\label{fig:histAm12_search} Distribution of \textit{restarted search} - Samples 2 and 3.}
\end{figure}

Figures \ref{fig:histAm12} and \ref{fig:histAm12_search}, also show a concentration of results on the left and a long tail on the right. 

\subsubsection{Empirical frequency distribution similarity tests}

The distributions were compared using the Kolmogorov-Smirnov (KS) test, which allowed the statistical evaluation of the similarity between pairs of the empirical frequency distributions as shown in Table \ref{tab:KS_test}(a) and (b). The SAAFB and HS distributions are statistically similar in all cases, while the RSFB algorithm distribution has no statistically significant similarity with any others.

\begin{table}[H]
\centering
\tiny
\caption{Distribution comparison with KS} 
\label{tab:KS_test}
 \subfloat[\textit{computational cost} metric.]{
 \centering
    \begin{tabular}{crrrc}
    \hline
    Sample & Comparison & statistic D & p value & Similar\\ 
    \hline
        1 & RSFB vs SAAFB & 0.17428 & 2.2e-16 & No \\ 
        1 & RSFB vs HS    & 0.17202 & 2.2e-16 & No \\ 
        1 & SAAFB vs HS   & 0.01049 & 0.9518  & Yes \\ 
        2 & SAAFB vs HS   & 0.00340 & 1.0000  & Yes \\ 
        3 & SAAFB vs HS   & 0.00228 & 1.0000  & Yes \\ 
        \hline
    \end{tabular}
 }
  \hspace{0.1cm} 
 \subfloat[\textit{restarted search} metric.]{
 \centering
    \begin{tabular}{crrrc}
    \hline
    Sample & Comparison & statistic D & p value & Similar\\ 
    \hline
    1 & RSFB vs SAAFB & 0.57613 & 2.2e-16 & No \\ 
    1 & RSFB vs HS    & 0.57613 & 2.2e-16 & No \\ 
    1 & SAAFB vs HS   & 0.00000 & 1.0000  & Yes \\ 
    2 & SAAFB vs HS   & 0.00000 & 1.0000  & Yes \\ 
    3 & SAAFB vs HS   & 0.00000 & 1.0000  & Yes \\ 
     \hline
    \end{tabular}
 }
\end{table}

\subsection{Correlation between factors and performance metrics}
Spearman's Coefficient correlation method was used due to the non-parametric nature of the data. The correlation took into account only the maximum result of each algorithm per factor of the experiment (in both metrics). In the case of the factor $f_5$ (\textit{percNeg}), in all algorithms, the correlation was evaluated only for values with up to 50\% of negative activities, as the algorithms reverse the search direction when this value is greater.\\

\begin{table}[!h]
\centering
\tiny
\caption{Spearman's Coefficient (\textit{computational cost} vs \textit{factors}) - Sample 1.}
\begin{tabular}{rrrrrrrr}
  \hline
 & \textit{vertices} & \textit{layer} & \textit{maxDegree} & \textit{discRate} & \textit{percNeg} & \textit{cpMult} & \textit{edges}\\ 
  \hline
RSFB & 0.93 & -0.20 & NA & 0.67 & 0.94 & NA & 0.78\\ 
  SAAFB & 0.96 & -0.18 & NA & 0.70 & 1.00& NA & 0.85\\ 
  HS & 0.96 & -0.18 & NA & 0.70 & 1.00& NA & 0.85\\ 
   \hline
\end{tabular}
\label{tab:corAm1}
\end{table}

\begin{table}[!h]
\centering
\tiny
\caption{Spearman's Coefficient (\textit{restarted search} vs \textit{factors}) - Sample 1.}
\begin{tabular}{rrrrrrrr}
  \hline
 & \textit{vertices} & \textit{layer} & \textit{maxDegree} & \textit{discRate} & \textit{percNeg} & \textit{cpMult} & \textit{edges}\\ 
  \hline
RSFB & 0.86 & -0.29 & NA & 0.79 & 0.94 & NA & 0.68\\ 
  SAAFB & 0.78 & -0.40 & NA & 0.44 & 0.71& NA & 0.56\\ 
  HS & 0.78 & -0.40 & NA & 0.44 & 0.71& NA & 0.56\\ 
   \hline
\end{tabular}
\label{tab:corAm1_search}
\end{table}

\begin{table}[H]
\centering
\tiny
\caption{Spearman's Coefficient (\textit{computational cost} vs \textit{factors}) - Sample 2.} 
\label{tab:corAm2}
\begin{tabular}{rrrrrrrr}
  \hline
 & \textit{vertices} & \textit{layer} & \textit{maxDegree} & \textit{discRate} & \textit{percNeg} & \textit{cpMult} & \textit{edges}\\ 
  \hline
  SAAFB & 0.93 & -0.13 & NA & 0.37 & 1.00& NA & 0.76\\ 
  HS & 0.93 & -0.13 & NA & 0.37 & 1.00& NA & 0.75\\ 
   \hline
\end{tabular}
\end{table}

\begin{table}[H]
\centering
\tiny
\caption{Spearman's Coefficient (\textit{restarted search cost} vs \textit{factors}) - Sample 2.} 
\label{tab:corAm2_search}
\begin{tabular}{rrrrrrrr}
  \hline
 & \textit{vertices} & \textit{layer} & \textit{maxDegree} & \textit{discRate} & \textit{percNeg} & \textit{cpMult} & \textit{edges}\\ 
  \hline
  SAAFB & 0.70 & -0.32 & NA & 0.58 & 0.39& NA & 0.43\\ 
  HS & 0.70 & -0.32 & NA & 0.58 & 0.39& NA & 0.43\\ 
   \hline
\end{tabular}
\end{table}

\begin{table}[H]
\centering
\tiny
\caption{Spearman's Coefficient (\textit{computational cost} vs \textit{factors}) - Sample 3.} 
\label{tab:corAm3}
\begin{tabular}{rrrrrrrr}
  \hline
 & \textit{vertices} & \textit{layer} & \textit{maxDegree} & \textit{discRate} & \textit{percNeg} & \textit{cpMult} & \textit{edges}\\ 
  \hline
  SAAFB & 0.93 & NA & 0.96 & 0.75 & 1.00 & NA & 0.53 \\ 
  HS & 0.93 & NA & 0.96 & 0.75 & 1.00 & NA & 0.53\\ 
   \hline
\end{tabular}
\end{table}

\begin{table}[H]
\centering
\tiny
\caption{Spearman's Coefficient (\textit{restarted search} vs \textit{factors}) - Sample 3.} 
\label{tab:corAm3_search}
\begin{tabular}{rrrrrrrr}
  \hline
 & \textit{vertices} & \textit{layer} & \textit{maxDegree} & \textit{discRate} & \textit{percNeg} & \textit{cpMult} & \textit{edges}\\ 
  \hline
  SAAFB & 0.85 & NA & 0.88 & 0.88 & 0.99 & NA & 0.36 \\ 
  HS & 0.85 & NA & 0.88 & 0.88 & 0.99 & NA & 0.36 \\ 
   \hline
\end{tabular}
\end{table}

\begin{table}[H]
\centering
\tiny
\caption{Spearman's Coefficient (\textit{computational cost} vs runtime).} 
\label{tab:corCostTime}
\begin{tabular}{crrr}
    \hline
     & Sample 1 & Sample 2 & Sample 3 \\  
  \hline
          RSFB & 0.90 & NA & NA \\ 
          SAAFB & 0.79 & 0.79 & 0.96 \\ 
          HS & 0.86 & 0.92 & 0.96 \\ 
   \hline
\end{tabular}
\end{table}

\subsection{Discussion on the preliminary data analysis }
\subsubsection{Computational cost and restarted search}
Table \ref{tab:summaryAm1}(a), referring to \textit{computational cost} in Sample 1, shows that the values for the three algorithms in the first quartile and also in the second quartile (median) are very close. However, in the third quartile, the value for the RSFB algorithm is almost double the respective values of SAAFB and HS. It is also worth noting that the maximum value of RSFB is about five times greater than the respective values of SAAFB and HS.\\

In Table, \ref{tab:summaryAm1}(b), referring to \textit{restarted search} in Sample 1, the median is three times greater than the value of SAAFB and HS as the mean. In the third quartile, the value of RS is five times greater than the others. The maximum value of RS is about twenty times higher than others. It is also possible to notice that the maximum value of RSFB is about five times greater than the respective values of SAAFB and HS.\\

At the same time, for the metric \textit{computational cost}, it can be seen that the highest results in Sample 3 are on the order of 10,000,000, while in Sample 2 the highest results are on the order of 60,000. It is important to remember that Samples 2 and 3 have graphs that range from 16 to 320 vertices but that the highest results of Sample 3 are about 166 times greater than those of Sample 2 with the same algorithms (SAAFB and HS). This big difference is related to the high number of edges in the graphs of Sample 3, which are complete bipartite digraphs (disregarding \textit{dummies}).

\subsubsection{Empirical distribution similarity tests}

Results show that SAAFB and HS empirical distributions are \emph{statistically similar} while the distribution of the RSFB algorithm is  \emph{statistically different} from the other two.

\subsubsection{ Metrics corrrelation with experimental factors }

\textcolor{black}{
According to Table \ref{tab:corAm1}, in Sample 1, the highest correlation coefficients of the metric \textit{computational cost} are with the factors $f_1$ (\textit{vertices} ), $f_5$ (\textit {percNeg}), and $f_7$ (\textit{edges}). For the metric \textit{restarted search} the highest correlation coefficients were with $f_1$ (\textit{vertices}), $f_5$ (\textit{percNeg}), $f_7$ (\textit{edges}), and $f_4 $ (\textit{discRate}). The other factors did not present relevant coefficients. Since the factors $f_3$ (\textit{maxDegree}) and $f_6$ (\textit{cpMult}), have sampling intervals with two values, they were not included in the analysis.\\
}

\textcolor{black}{
In Sample 2, as shown in Table \ref{tab:corAm2}, the two factors with the highest correlation coefficients with the metric \textit{computational cost} were also $f_1$ (\textit{vertices}) and $f_5$ (\textit{percNeg}), both with values above 90\%. The other factors did not present relevant correlation coefficients. Table \ref{tab:corAm2_search} shows that the highest correlation of \textit{restarted search} was with $f_1$ $(vertices)$.\\
}
\textcolor{black}{
In Sample 3, according to Tables \ref{tab:corAm3} and \ref{tab:corAm3_search}, four factors showed strong correlation coefficients: $f_1$ (\textit{vertices}), $f_3$ (\textit {maxDegree}), $f_5$ (\textit{percNeg}), and $f_4$ ($discRate$). In this sample, unlike the first two, the factor $f_5$ (\textit{maxDegree}) includes a range of values between $(16-2)/2$ and $(320-2)/2$ (according to Table \ref{tab:lotes}). Therefore, the correlation coefficient, in this case, presented a relevant value. The other factors did not present relevant values.\\
}

\textcolor{black}{ 
Table \ref{tab:corCostTime} shows the correlation between the metric \textit{computational cost} with the  execution time of the algorithms (time expressed in milliseconds, runtime). As the algorithms can reverse the search, the values considered for the factor $f_5$ (\emph{percNeg})  were up to 50\%. Since the RSFB algorithm was not used in Samples 2 and 3, NA refers to not applicable. In the other cases, there is a strong ($\geq$ 70\%) or very strong ($\geq$ 90\%) correlation between the \emph{computational cost} and the time measure (which was not considered a dependent variable). No correlation study was made between \textit{restarted search} and execution time, as this metric does not refer to the total cost, as is the case with \textit{computational cost}.
}

\section{Results: statistical models}
\textcolor{black}{
The experimental results were analyzed with the Generalized Linear Model (GLM) statistical modeling technique. Twenty-eight statistical models were obtained from results of the two metrics (\emph{computational cost} and \emph{restarted search}). All the models assumed a Poisson-Gamma distribution (negative binomial), which agrees with the results from the preliminary analysis. Two model types were developed, with one factor and two factors, considering the $maxCost$ formulation of the subsection \ref{sec:max-cost}.
}
\subsection{Models of Sample 1}
\textcolor{black}{
The results from Sample 1 (networks between 16 and 80 vertices) enabled the creation of eleven models: nine referring to the \emph{computational cost} metric and three referring to the \emph{restarted search} metric.
}
\subsubsection{Models for computational cost (Sample 1)}

\begin{figure}[H]
\centering
    \fbox{a.\includegraphics[scale=0.24]{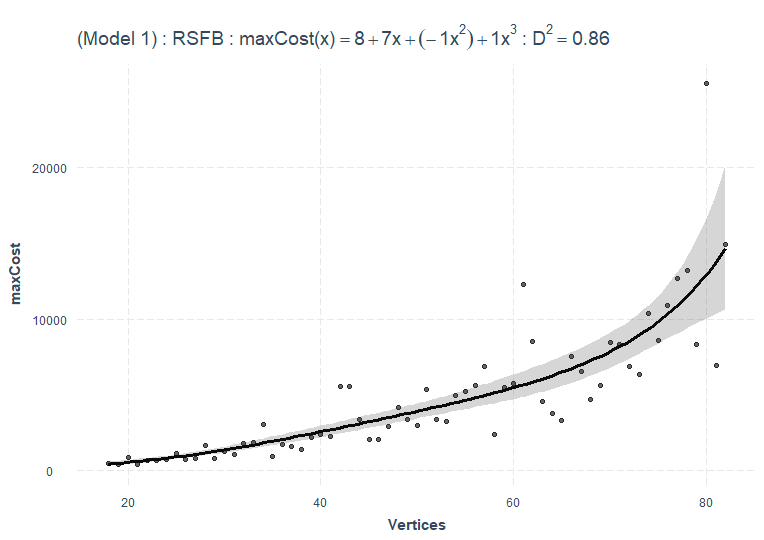}}   
    \hspace{10px}
    \fbox{b.\includegraphics[scale=0.24]{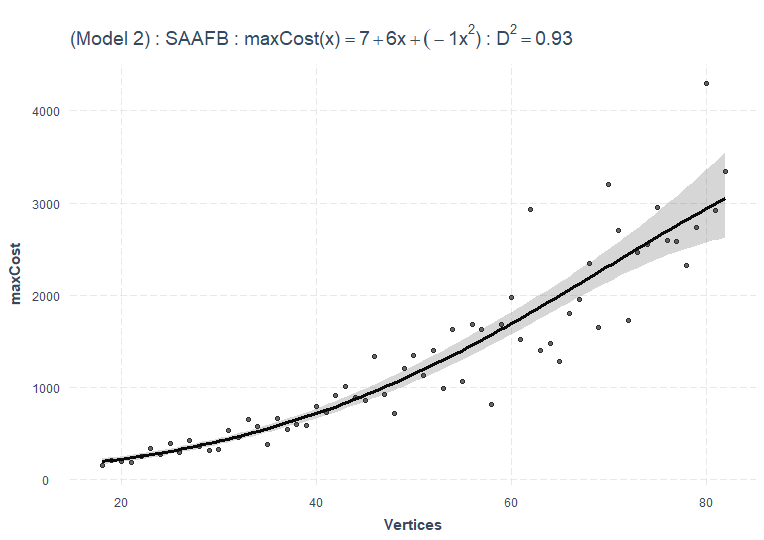}}
    \caption{RSFB and SAAFB $maxCost(vertices)$ for \emph{computational cost} - Sample 1.}
    \label{fig:sample1a_one_factor}
\end{figure}

\textcolor{black}{
Model 1 refers to $maxCost(vertices)$ for the RSFB algorithm as a function of factor $vertices$. Figure \ref{fig:sample1a_one_factor}(a) highlights the curve, polynomial, and the proportion of deviation explained with $D^2$ (GLM equivalent to $R^2$). Table \ref{tab:model1} shows the respective estimated value, standard error, value of Wald's $z$ statistic and the value $Pr(>|z|)$. In Model 1, the polynomial is: $maxCost(x) = 8 + 7x -1x^2 + 1x^3$, where $x$ is $vertices$.\\}

\begin{table}[ht]
\centering
\tiny
\caption{RSFB $maxCost(vertices)$ for \emph{computational cost} - Sample 1.}
\label{tab:model1}
\begin{tabular}{rrrrr}
  \hline
(Model 1) & Estimate & Std. Error & z value & Pr($>$$|$z$|$) \\ 
  \hline
(Intercept) & 8.1181 & 0.0425 & 191.19 & 0.0000 \\ 
  poly(x, 3)1 & 7.2567 & 0.3427 & 21.18 & 0.0000 \\ 
  poly(x, 3)2 & -1.1065 & 0.3426 & -3.23 & 0.0012 \\ 
  poly(x, 3)3 & 0.7793 & 0.3426 & 2.27 & 0.0229 \\ 
   \hline
\end{tabular}

\end{table}

\begin{figure}[H]
\centering
    \fbox{a.\includegraphics[scale=0.24]{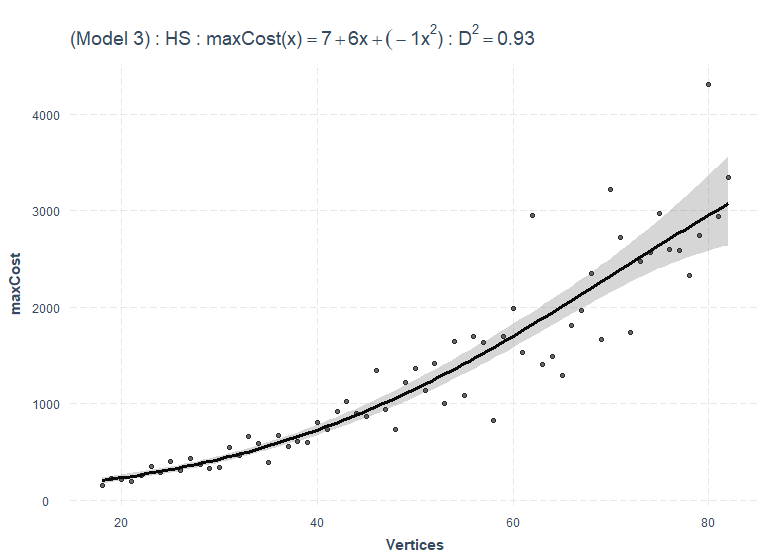}}
    \hspace{10px}
    \fbox{b.\includegraphics[scale=0.24]{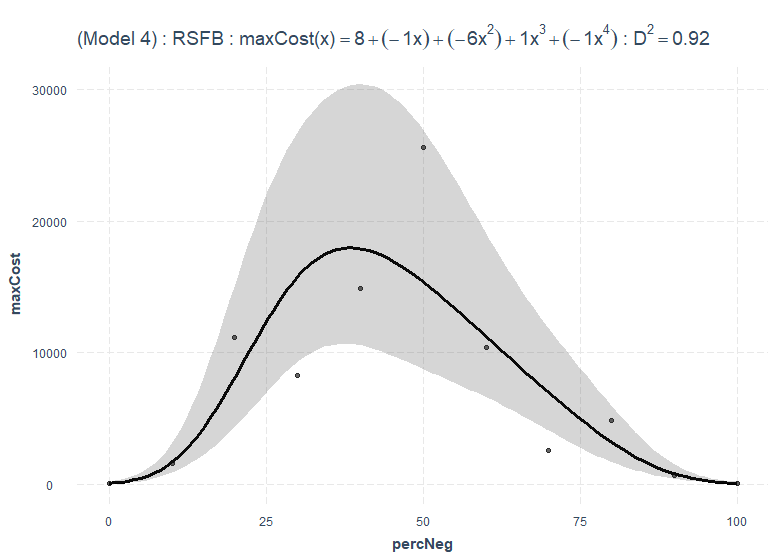}}
    \caption{HS $maxCost(vertices)$, RSFB $maxCost(percNeg)$ for \emph{computational cost} - Sample 1.}
    \label{fig:sample1b_one_factor}
\end{figure}

\textcolor{black}{
Models 2 and 3 refer to the $maxCost(vertices)$ of the SAAFB and HS algorithms. Figure \ref{fig:sample1a_one_factor}(b) and Figure \ref{fig:sample1b_one_factor}(a) highlight the respective curves, polynomials and $D^2$ of the models. Tables \ref{tab:models2_3}(a) and \ref{tab:models2_3}(b) show the results of the models. Models 2 and 3 have the same polynomial: $maxCost(x) = 7 + 6x - 1x^2$, where $x$ refers to the $vertices$.
}

\begin{table}[H]
\centering
\tiny
\caption{SAAFB and HS $maxCost(vertices)$ for \emph{computational cost} - Sample 1.}
\label{tab:models2_3}
 \subfloat[Model 2 - SAAFB]{
 \centering
    \begin{tabular}{rrrrr}
    \hline
    (Model 2) & Estimate & Std. Error & z value & Pr($>$$|$z$|$) \\ 
    \hline
    (Intercept) & 6.9091 & 0.0261 & 264.79 & 0.0000 \\ 
    poly(x, 2)1 & 6.4801 & 0.2115 & 30.64 & 0.0000 \\ 
    poly(x, 2)2 & -0.9662 & 0.2112 & -4.57 & 0.0000 \\ 
    \hline
    \end{tabular}
    
 }
 \hspace{0.1cm} 
 \subfloat[Model 3 - HS]{
 \centering
    \begin{tabular}{rrrrr}
      \hline
      (Model 3) & Estimate & Std. Error & z value & Pr($>$$|$z$|$) \\ 
      \hline
      (Intercept) & 6.9174 & 0.0260 & 265.80 & 0.0000 \\ 
      poly(x, 2)1 & 6.4379 & 0.2109 & 30.53 & 0.0000 \\ 
      poly(x, 2)2 & -0.9455 & 0.2106 & -4.49 & 0.0000 \\ 
      \hline
    \end{tabular}
 }
\end{table}

\begin{figure}[H]
\centering
    \fbox{a.\includegraphics[scale=0.24]{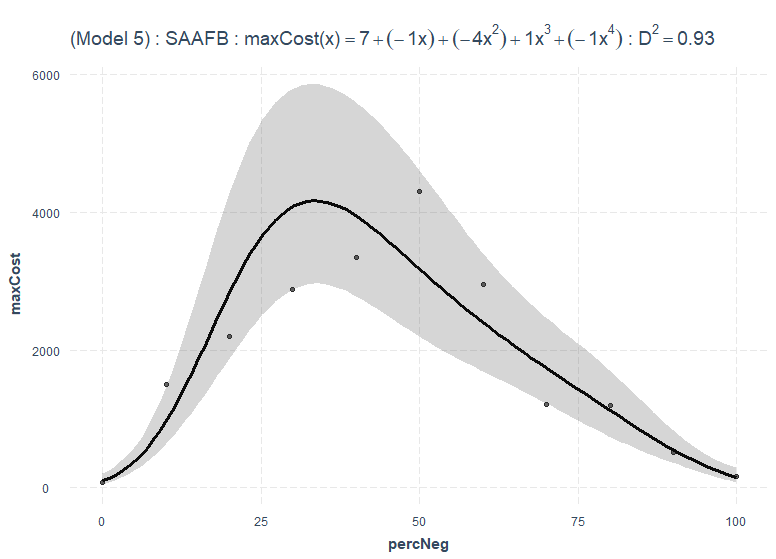}}
    \hspace{10px}
    \fbox{b.\includegraphics[scale=0.24]{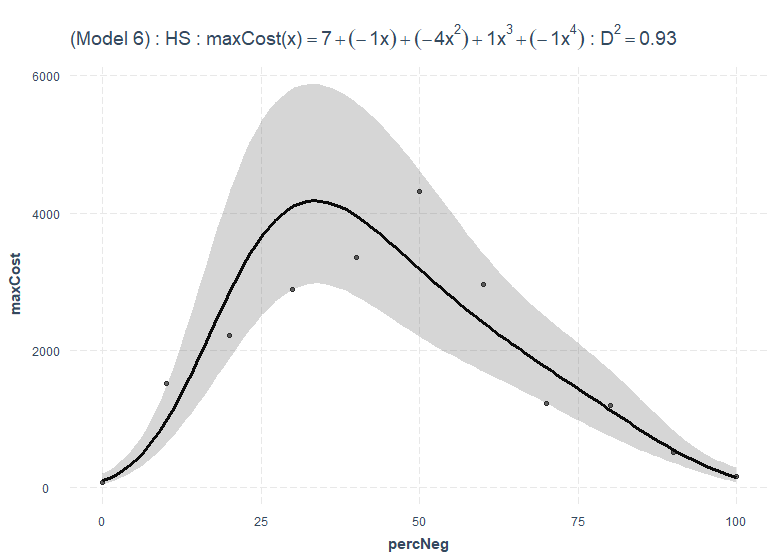}} 
    \caption{SAAFB and HS $maxCost(percNeg)$ for \emph{computational cost} - Sample 1.}
    \label{fig:sample1c_one_factor}
\end{figure}

\textcolor{black}{Models 4, 5, and 6 refer to the $maxCost(percNeg)$ of the RSFB, SAAFB, and HS algorithms for the \emph{computational cost}. Figures \ref{fig:sample1b_one_factor}(b), \ref{fig:sample1c_one_factor}(a), and \ref{fig:sample1c_one_factor}(b) highlight the respective curves, polynomials, and $D^2$. Tables \ref{tab:model4}, \ref{tab:models5_6}(a), and \ref{tab:models5_6}(b) display the model results. The polynomial of Model 4 is: $maxCost(x) = 8 - 6x^2 - 1x^4$, where $x$ is $percNeg$. Models 5 and 6 have the same polynomial: $ maxCost(x) = 7 - 4x^2 + 1x^3 - 1x^4$, where $x$ is $percNeg$.
}

\begin{table}[ht]
\centering
\caption{RSFB $maxCost(percNeg)$ for \emph{computational cost} - Sample 1.}
\tiny
\begin{tabular}{rrrrr}
  \hline
 (Model 4) & Estimate & Std. Error & z value & Pr($>$$|$z$|$) \\ 
  \hline
(Intercept) & 7.9374 & 0.1205 & 65.87 & 0.0000 \\ 
  poly(x, 4)1 & -0.7447 & 0.4038 & -1.84 & 0.0652 \\ 
  poly(x, 4)2 & -6.1344 & 0.4047 & -15.16 & 0.0000 \\ 
  poly(x, 4)3 & 0.7539 & 0.4030 & 1.87 & 0.0614 \\ 
  poly(x, 4)4 & -1.0992 & 0.4008 & -2.74 & 0.0061 \\ 
   \hline
\end{tabular}
    \label{tab:model4}
\end{table}

\begin{table}[ht]
\centering
\tiny
\caption{SAAFB and HS $maxCost(percNeg)$ for \emph{computational cost} - Sample 1.}
\label{tab:models5_6}
 \subfloat[Model 5 - SAAFB]{
 \centering
    \begin{tabular}{rrrrr}
      \hline
     (Model 5)& Estimate & Std. Error & z value & Pr($>$$|$z$|$) \\ 
      \hline
    (Intercept) & 7.0701 & 0.0799 & 88.45 & 0.0000 \\ 
      poly(x, 4)1 & -0.5058 & 0.2691 & -1.88 & 0.0602 \\ 
      poly(x, 4)2 & -3.7575 & 0.2696 & -13.94 & 0.0000 \\ 
      poly(x, 4)3 & 0.9398 & 0.2681 & 3.51 & 0.0005 \\ 
      poly(x, 4)4 & -0.8128 & 0.2662 & -3.05 & 0.0023 \\ 
       \hline
    \end{tabular}
    
 }
 \hspace{0.1cm} 
 \subfloat[Model 6 - HS]{
 \centering
    \begin{tabular}{rrrrr}
      \hline
     (Model 6)& Estimate & Std. Error & z value & Pr($>$$|$z$|$) \\ 
      \hline
    (Intercept) & 7.0762 & 0.0799 & 88.53 & 0.0000 \\ 
      poly(x, 4)1 & -0.5035 & 0.2691 & -1.87 & 0.0613 \\ 
      poly(x, 4)2 & -3.7462 & 0.2696 & -13.90 & 0.0000 \\ 
      poly(x, 4)3 & 0.9379 & 0.2681 & 3.50 & 0.0005 \\ 
      poly(x, 4)4 & -0.8118 & 0.2662 & -3.05 & 0.0023 \\ 
       \hline
    \end{tabular}
 }
\end{table}

\begin{figure}[H]
\centering
    \fbox{a.\includegraphics[scale=0.28]{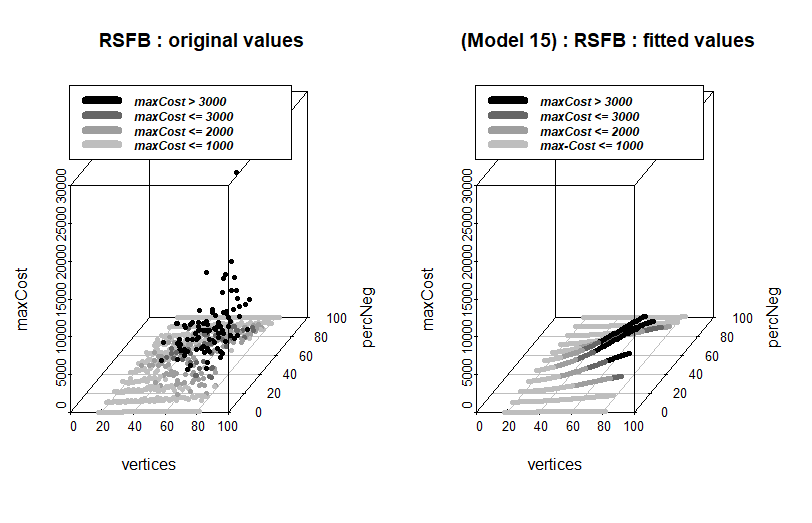}}
    \hspace{5px}
    \fbox{b.\includegraphics[scale=0.28]{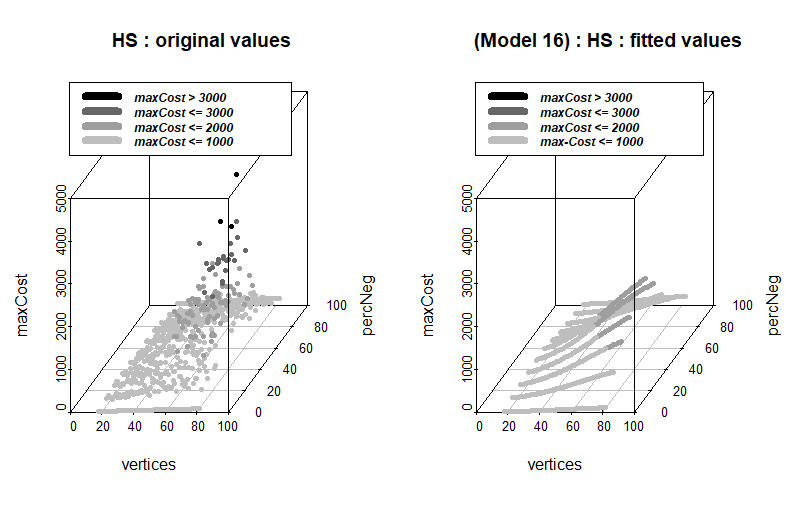}}
    \hspace{5px}
    \begin{center}\fbox{c\includegraphics[scale=0.28]{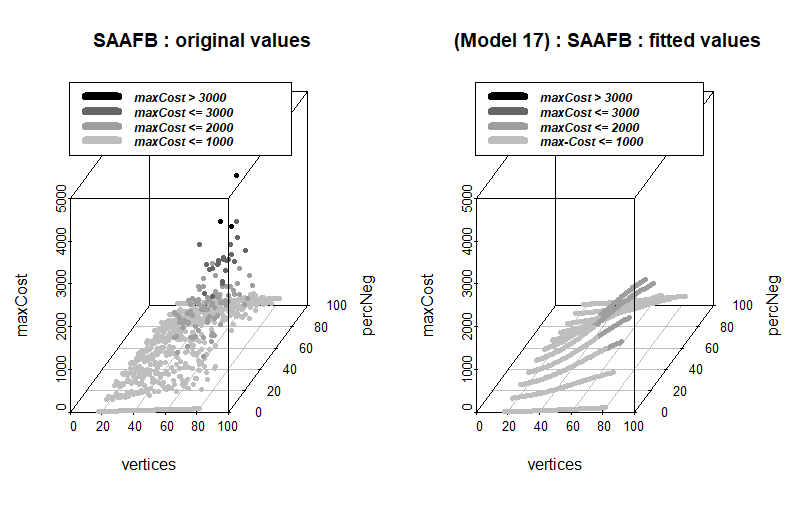}}\end{center}
    \caption{Function $maxCost(vertices, percNeg)$ for \emph{computational cost} - Sample 1.}
    \label{fig:sample1_two_factor}
\end{figure}

\textcolor{black}{
In addition to the one-factor models, three two-factor models were developed for \emph{computational cost}. Figures \ref{fig:sample1_two_factor}(a), \ref{fig:sample1_two_factor}(b), and \ref{fig:sample1_two_factor}(c) highlight each of these models with their original data. The axes of the models refer to $x (vertices)$, $z (percNeg)$ and $y (maxCost)$. Table \ref{tab:model15} refers to the first model with two factors, i.e., $maxCost(vertices, percNeg)$ for the RSFB. In Model 15, the polynomial is: $maxCost(x, y) = 6 + 19x -3y -3x^2 -35y^2 + 4y^3 - 3y^4 - 3y^6$, where $x$ refers to the $vertices$ and $y$ the $percNeg$. $D^2$ of Model 15 was 88\%.
}

\begin{table}[H]
\centering
\caption{RSFB $maxCost(vertices, percNeg)$ for \emph{computational cost} - Sample 1.}
\tiny
\begin{tabular}{rrrrr}
  \hline
 (Model 15)& Estimate & Std. Error & z value & Pr($>$$|$z$|$) \\ 
  \hline
(Intercept) & 6.2855 & 0.0180 & 349.40 & 0.0000 \\ 
  poly(x, 2)1 & 19.3530 & 0.4824 & 40.12 & 0.0000 \\ 
  poly(x, 2)2 & -3.0407 & 0.4818 & -6.31 & 0.0000 \\ 
  poly(y, 6)1 & -3.3526 & 0.4885 & -6.86 & 0.0000 \\ 
  poly(y, 6)2 & -35.2969 & 0.4894 & -72.12 & 0.0000 \\ 
  poly(y, 6)3 & 3.5723 & 0.4867 & 7.34 & 0.0000 \\ 
  poly(y, 6)4 & -3.0889 & 0.4831 & -6.39 & 0.0000 \\ 
  poly(y, 6)5 & -0.2048 & 0.4800 & -0.43 & 0.6696 \\ 
  poly(y, 6)6 & -3.0378 & 0.4780 & -6.35 & 0.0000 \\ 
   \hline
\end{tabular}
    \label{tab:model15}
\end{table}

\begin{table}[ht]
\centering
\tiny
\caption{HS and SAAFB $maxCost(vertices, percNeg)$ for \emph{computational cost} - Sample 1.}
\label{tab:models16_17}
 \subfloat[Model 16 - HS]{
 \centering
    \begin{tabular}{rrrrr}
      \hline
     (Model 16)& Estimate & Std. Error & z value & Pr($>$$|$z$|$) \\ 
      \hline
    (Intercept) & 5.7844 & 0.0127 & 455.76 & 0.0000 \\ 
      poly(x, 2)1 & 16.4096 & 0.3407 & 48.17 & 0.0000 \\ 
      poly(x, 2)2 & -2.7184 & 0.3400 & -7.99 & 0.0000 \\ 
      poly(y, 6)1 & -0.6613 & 0.3463 & -1.91 & 0.0562 \\ 
      poly(y, 6)2 & -22.2871 & 0.3467 & -64.29 & 0.0000 \\ 
      poly(y, 6)3 & 5.8396 & 0.3444 & 16.95 & 0.0000 \\ 
      poly(y, 6)4 & -3.1664 & 0.3413 & -9.28 & 0.0000 \\ 
      poly(y, 6)5 & 1.8862 & 0.3386 & 5.57 & 0.0000 \\ 
      poly(y, 6)6 & -2.4509 & 0.3368 & -7.28 & 0.0000 \\
       \hline
    \end{tabular}
 }
 \hspace{0.1cm} 
 \subfloat[Model 17 - SAAFB]{
 \centering
    \begin{tabular}{rrrrr}
      \hline
     (Model 17)& Estimate & Std. Error & z value & Pr($>$$|$z$|$) \\ 
      \hline
    (Intercept) & 5.7697 & 0.0127 & 453.51 & 0.0000 \\ 
      poly(x, 2)1 & 16.5993 & 0.3415 & 48.60 & 0.0000 \\ 
      poly(x, 2)2 & -2.7960 & 0.3409 & -8.20 & 0.0000 \\ 
      poly(y, 6)1 & -0.6749 & 0.3473 & -1.94 & 0.0520 \\ 
      poly(y, 6)2 & -22.4429 & 0.3477 & -64.55 & 0.0000 \\ 
      poly(y, 6)3 & 5.8525 & 0.3454 & 16.95 & 0.0000 \\ 
      poly(y, 6)4 & -3.1730 & 0.3421 & -9.27 & 0.0000 \\ 
      poly(y, 6)5 & 1.8941 & 0.3394 & 5.58 & 0.0000 \\ 
      poly(y, 6)6 & -2.4553 & 0.3375 & -7.27 & 0.0000 \\ 
       \hline
    \end{tabular}
 }
\end{table}

Tables \ref{tab:models16_17}(a) and \ref{tab:models16_17}(b) refere to Models 16 and 17. These models have the same polynomial: $maxCost(x) = 6 + 16x - 1y - 3x^2 - 22y^2 + 6y^3 - 3y^4 + 2y^5 + 2y^6$, where $x$ is $vertices$ and $y$ is $percNeg$. $D^2$ of Models 16 and 17 were 89\%.

\subsubsection{Models for restarted search (Sample 1)}

\begin{figure}[H]
\centering
    \fbox{\includegraphics[scale=0.24]{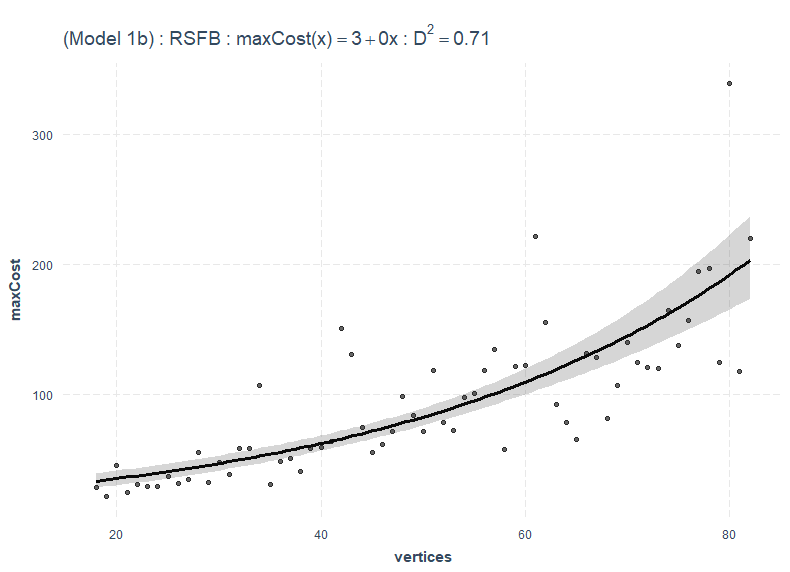}}
    \caption{RSFB $maxCost(vertices)$ for \emph{restarted search} - Sample 1.}
    \label{fig:model1b}
\end{figure}

\begin{figure}[H]
\centering
    \fbox{a\includegraphics[scale=0.24]{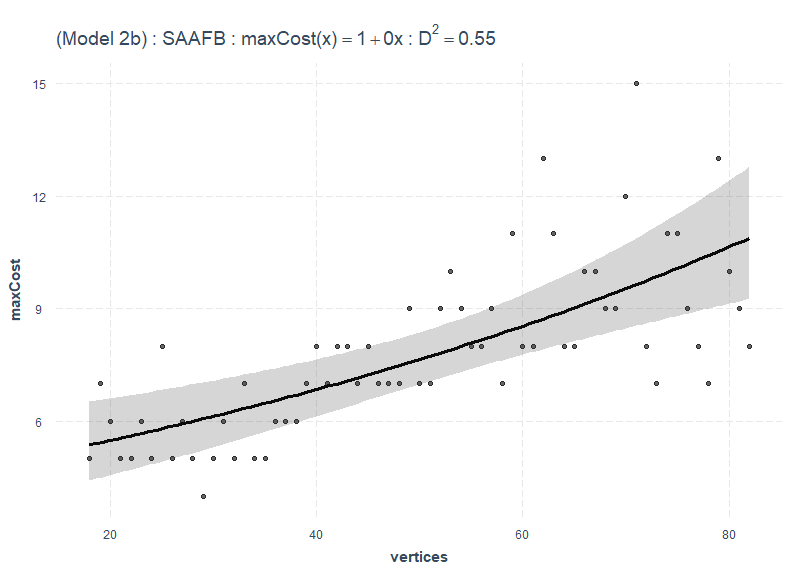}}
    \hspace{5px}
    \fbox{b\includegraphics[scale=0.24]{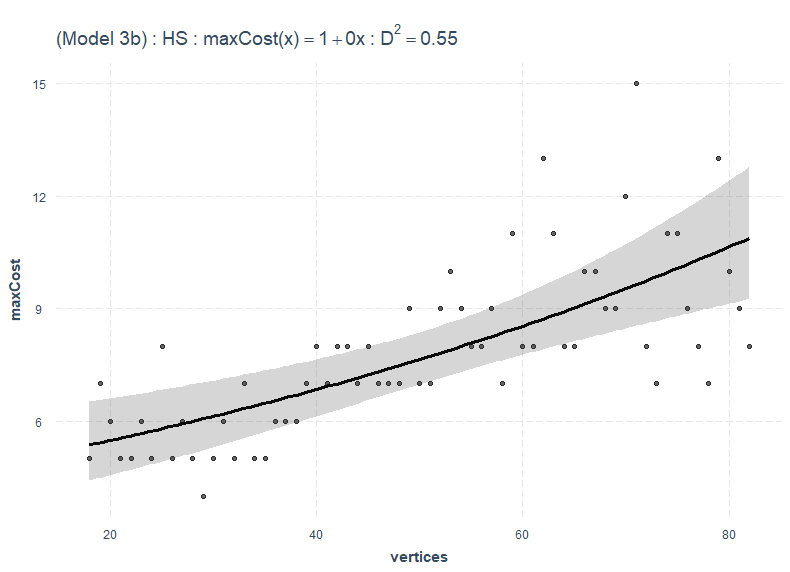}}
    \caption{SAAFB and HS $maxCost(vertices)$ for \emph{restarted search} - Sample 1.}
    \label{fig:model2b_3b}
\end{figure}

 \textcolor{black}{
For the \emph{restarted search} metric, in Sample 1, three models were built. Models 1b, 2b, and 3b refer to the $maxCost(vertices)$ of the RSFB, SAAFB, and HS algorithms, as shown in Figures \ref{fig:model1b} and \ref{fig:model2b_3b}. Table \ref{tab:model1b} shows that the polynomial of Model 1b is: $maxCost(x) = 3 + 0x$, where $x$ is $vertices$. Tables \ref{tab:models2b_3b}(a) and \ref{tab:models2b_3b}(b) show that the Models 2b and 3b have the same polynomial: $maxCost(x) = 1 + 0x$, where $x$ is $vertices$.
}

\begin{table}[H]
\centering
\tiny
\caption{RSFB $maxCost(vertices)$ for \emph{restarted search} - Sample 1.}
\label{tab:model1b}
\begin{tabular}{rrrrr}
  \hline
 (Model 1b) & Estimate & Std. Error & z value & Pr($>$$|$z$|$) \\ 
  \hline
(Intercept) & 3.0165 & 0.1177 & 25.63 & 0.0000 \\ 
  x & 0.0280 & 0.0022 & 12.97 & 0.0000 \\ 
   \hline
\end{tabular}
\end{table}

\begin{table}[H]
\centering
\tiny
\caption{SAAFB and HS $maxCost(vertices)$ for \emph{restarted search} - Sample 1.}
\label{tab:models2b_3b}
 \subfloat[Model 2b - SAAFB]{
 \centering
 \begin{tabular}{rrrrr}
  \hline
 (Model 2b)& Estimate & Std. Error & z value & Pr($>$$|$z$|$) \\ 
  \hline
  (Intercept) & 1.4832 & 0.1364 & 10.87 & 0.0000 \\ 
  x & 0.0110 & 0.0024 & 4.61 & 0.0000 \\ 
  \hline
\end{tabular}
    
 }
 \hspace{0.1cm} 
 \subfloat[Model 3b - HS]{
 \centering
 \begin{tabular}{rrrrr}
  \hline
 (Model 3b)& Estimate & Std. Error & z value & Pr($>$$|$z$|$) \\ 
  \hline
 (Intercept) & 1.4832 & 0.1364 & 10.87 & 0.0000 \\ 
  x & 0.0110 & 0.0024 & 4.61 & 0.0000 \\ 
   \hline
\end{tabular}
 }
\end{table}

\textcolor{black}{
It is worth noting that models for the factor $edges$ were also obtained for RSFB, SAAFB, and HS. However, their results were similar to models with the factor $vertices$ due to the proportionality between $vertices$ and $edges$ in data from Sample 1.
}

\subsection{Models of Sample 2}

\textcolor{black}{
The results from Sample 2 (networks between 16 and 320 vertices) enabled the creation of eight models: six referring to the \emph{computational cost} and two referring to the \emph{restarted search}.
}

\subsubsection{Models for computational cost (Sample 2)}

\begin{figure}[H]
\centering
    
    \fbox{a.\includegraphics[scale=0.24]{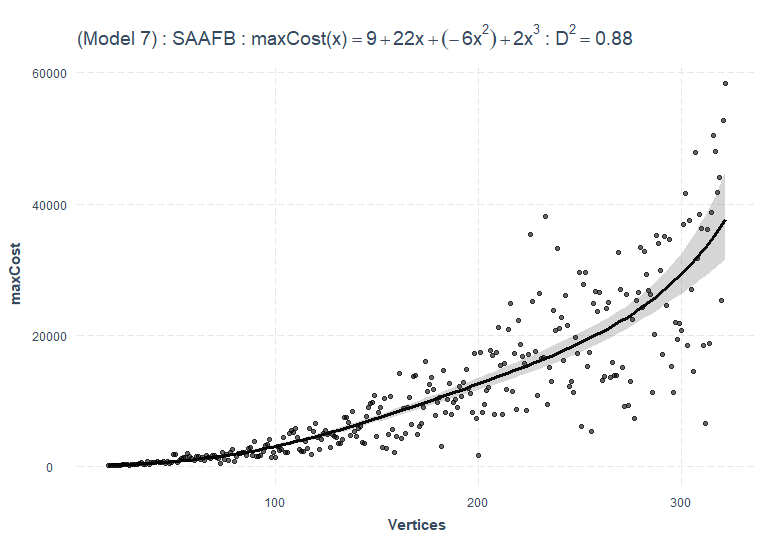}}   
    \hspace{10px}
    \fbox{b.\includegraphics[scale=0.24]{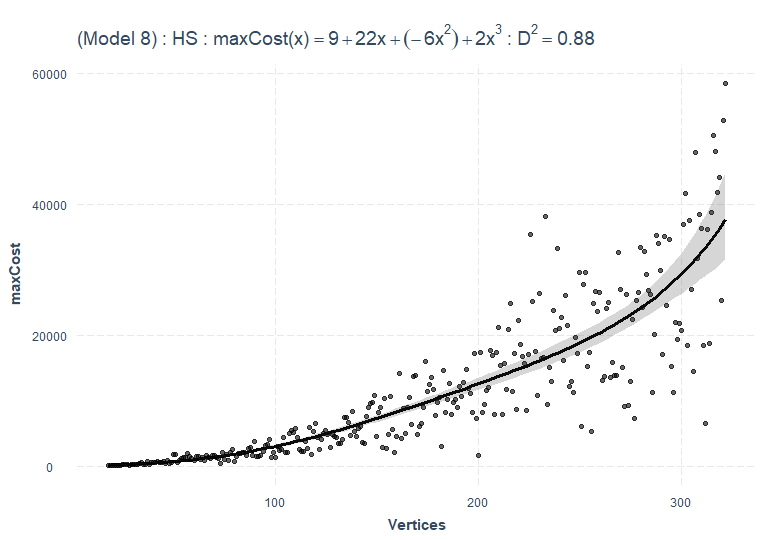}}
    \caption{SAAFB, HS $maxCost(vertices)$ for \emph{computational cost} - Sample 2.}
    \label{fig:sample2a_one_factor}
\end{figure}

\textcolor{black}{
Models 7 and 8 refer to the $maxCost(vertices)$ of the SAAFB and HS algorithms for the \emph{computational cost} metric, as shown in Figures \ref{fig:sample2a_one_factor}(a) and \ref{fig:sample2a_one_factor}(b). Tables \ref{tab:models7_8}(a) and \ref{tab:models7_8}(b) show that the models have the same polynomial:  $maxCost (x) = 9 + 22x - 6x^2 + 2x^3$, where $x$ is $vertices$.
}

\begin{table}[H]
\centering
\tiny
\caption{SAAFB, HS $maxCost(vertices)$ for \emph{computational cost} - Sample 2.}
\label{tab:models7_8}
 \subfloat[Model 7 - SAAFB]{
 \centering
    \begin{tabular}{rrrrr}
      \hline
     (Model 7)& Estimate & Std. Error & z value & Pr($>$$|$z$|$) \\ 
      \hline
    (Intercept) & 8.7534 & 0.0228 & 383.93 & 0.0000 \\ 
      poly(x, 3)1 & 22.4161 & 0.3987 & 56.23 & 0.0000 \\ 
      poly(x, 3)2 & -6.0673 & 0.3987 & -15.22 & 0.0000 \\ 
      poly(x, 3)3 & 2.2616 & 0.3987 & 5.67 & 0.0000 \\ 
       \hline
    \end{tabular}
 }
 \hspace{0.1cm} 
 \subfloat[Model 8 - HS]{
 \centering
\begin{tabular}{rrrrr}
  \hline
     (Model 8)& Estimate & Std. Error & z value & Pr($>$$|$z$|$) \\ 
      \hline
    (Intercept) & 8.7560 & 0.0228 & 384.63 & 0.0000 \\ 
      poly(x, 3)1 & 22.3701 & 0.3981 & 56.20 & 0.0000 \\ 
      poly(x, 3)2 & -6.0328 & 0.3981 & -15.15 & 0.0000 \\ 
      poly(x, 3)3 & 2.2387 & 0.3981 & 5.62 & 0.0000 \\ 
       \hline
    \end{tabular}
 }
\end{table}

\begin{figure}[H]
\centering
    \fbox{a.\includegraphics[scale=0.24]{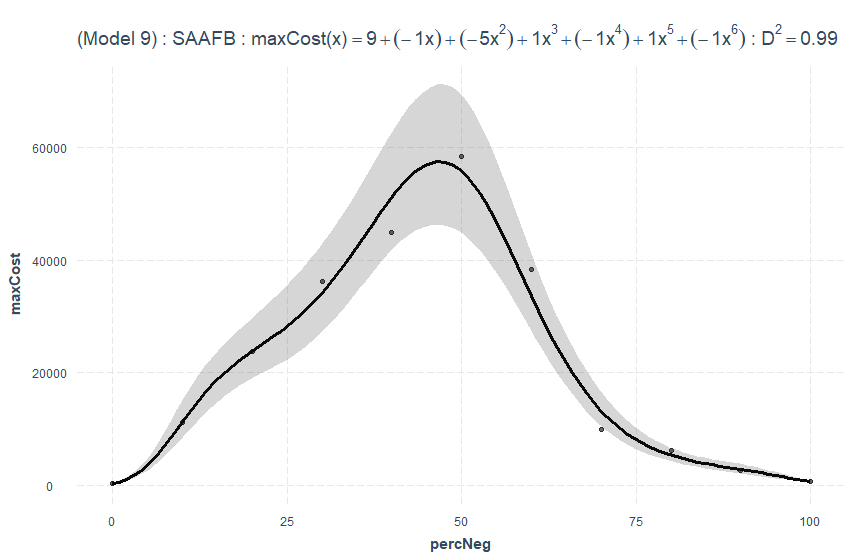}}
    \hspace{10px}
    \fbox{b.\includegraphics[scale=0.24]{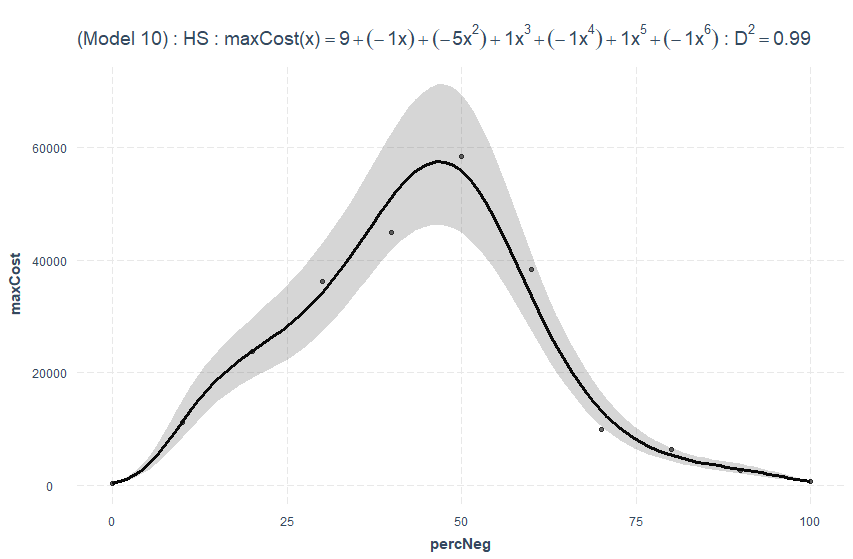}}   
    \caption{SAAFB, HS $maxCost(percNeg)$ for \emph{computational cost} - Sample 2.}
    \label{fig:sample2b_one_factor}
\end{figure}

\textcolor{black}{
Models 9 and 10 refer to the $maxCost(percNeg)$ of the SAAFB and HS algorithms, as shown in Figures \ref{fig:sample2b_one_factor}(a) and \ref{fig:sample2b_one_factor}(b). Tables \ref{tab:models9_10}(a) and \ref{tab:models9_10}(b) highlight that the models have the same polynomial: $maxCost(x) = 9 - 1x - 5x^2 + x^3 - 1x^4 + 1x^5 - 1x^6$, where $x$ is $percNeg$.
}
\begin{table}[H]
\centering
\tiny
\caption{SAAFB, HS $maxCost(percNeg)$ for \emph{computational cost} - Sample 2.}
\label{tab:models9_10}
 \subfloat[Model 9 - SAAFB]{
 \centering
    \begin{tabular}{rrrrr}
      \hline
     (Model 9)& Estimate & Std. Error & z value & Pr($>$$|$z$|$) \\
      \hline
    (Intercept) & 9.1172 & 0.0349 & 260.94 & 0.0000 \\ 
      poly(x, 6)1 & -0.8530 & 0.1187 & -7.19 & 0.0000 \\ 
      poly(x, 6)2 & -5.1860 & 0.1191 & -43.54 & 0.0000 \\ 
      poly(x, 6)3 & 1.3668 & 0.1181 & 11.57 & 0.0000 \\ 
      poly(x, 6)4 & -0.6875 & 0.1166 & -5.89 & 0.0000 \\ 
      poly(x, 6)5 & 0.5134 & 0.1154 & 4.45 & 0.0000 \\ 
      poly(x, 6)6 & -0.7549 & 0.1146 & -6.58 & 0.0000 \\ 
       \hline
    \end{tabular}
 }
 \hspace{0.1cm} 
 \subfloat[Model 10 - HS]{
 \centering
    \begin{tabular}{rrrrr}
      \hline
     (Model 10)& Estimate & Std. Error & z value & Pr($>$$|$z$|$) \\ 
      \hline
    (Intercept) & 9.1186 & 0.0349 & 261.04 & 0.0000 \\ 
      poly(x, 6)1 & -0.8529 & 0.1186 & -7.19 & 0.0000 \\ 
      poly(x, 6)2 & -5.1829 & 0.1191 & -43.53 & 0.0000 \\ 
      poly(x, 6)3 & 1.3665 & 0.1181 & 11.57 & 0.0000 \\ 
      poly(x, 6)4 & -0.6871 & 0.1166 & -5.89 & 0.0000 \\ 
      poly(x, 6)5 & 0.5138 & 0.1154 & 4.45 & 0.0000 \\ 
      poly(x, 6)6 & -0.7546 & 0.1146 & -6.58 & 0.0000 \\ 
       \hline
    \end{tabular}
 }
\end{table}

\begin{figure}[H] 
\centering
    \fbox{a.\includegraphics[scale=0.24]{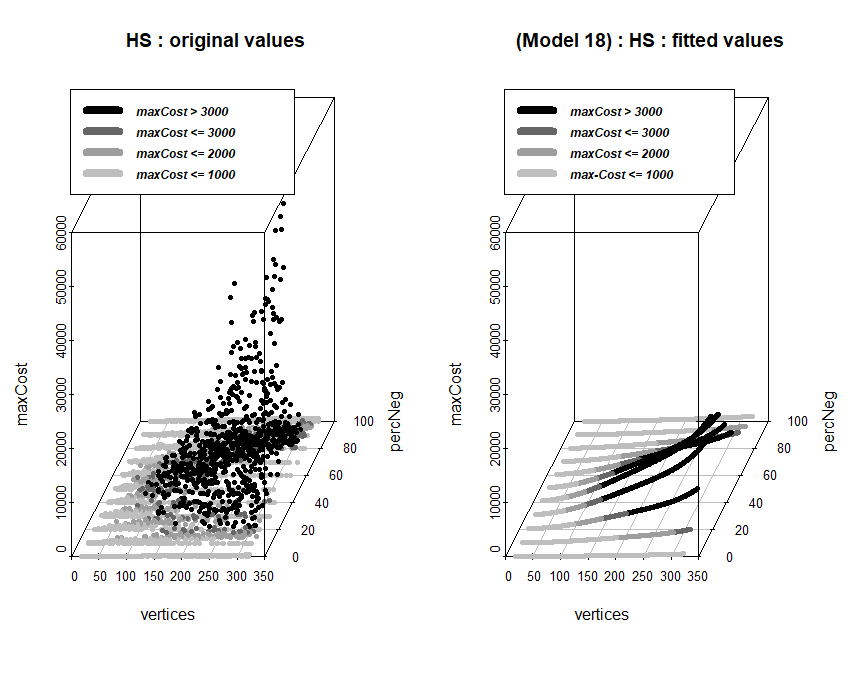}}
    \hspace{5px}
    \fbox{b.\includegraphics[scale=0.24]{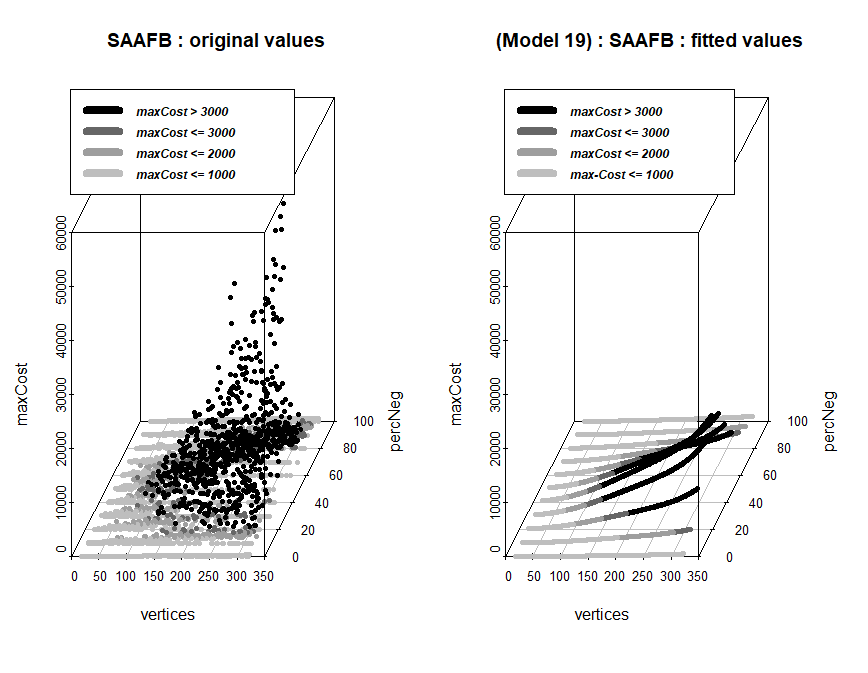}}

    \caption{HS and SAAFB $maxCost(vertices, percNeg)$ for \emph{computational} - Sample 2.}
    \label{fig:sample2c_two_factor}
\end{figure}

\textcolor{black}{
With two factors, Models 18 and 19 refer to the $maxCost(vertices, percNeg)$ of the HS and SAAFB algorithms, as shown in Figures \ref{fig:sample2c_two_factor}(a) and \ref{fig:sample2c_two_factor}(b). Tables \ref{tab:models18_19}(a) and \ref{tab:models18_19}(b) show that the models have the same polynomial: $maxCost(x) = 7 + 58x - 6y - 15x^2 - 61y^2 + 5x^3 + 19y^3 - 1y^4 + 1y^5 - 6y^6$, where $x$ is $vertices$ and $y$ is $percNeg$. Models 18 and 19 present the same $D^2$, 42\%.
}

\begin{table}[H]
\centering
\tiny
\caption{HS and SAAFB $maxCost(vertices, percNeg)$ for \emph{computational} - Sample 2.}
\label{tab:models18_19}
 \subfloat[Model 18 - HS]{
 \centering
    \begin{tabular}{rrrrr}
      \hline
     (Model 18)& Estimate & Std. Error & z value & Pr($>$$|$z$|$) \\ 
      \hline
    (Intercept) & 6.9367 & 0.0262 & 265.07 & 0.0000 \\ 
      poly(x, 3)1 & 57.9835 & 1.5165 & 38.24 & 0.0000 \\ 
      poly(x, 3)2 & -14.8337 & 1.5165 & -9.78 & 0.0000 \\ 
      poly(x, 3)3 & 5.1964 & 1.5164 & 3.43 & 0.0006 \\ 
      poly(y, 6)1 & -5.7769 & 1.5168 & -3.81 & 0.0001 \\ 
      poly(y, 6)2 & -60.6295 & 1.5169 & -39.97 & 0.0000 \\ 
      poly(y, 6)3 & 18.8305 & 1.5165 & 12.42 & 0.0000 \\ 
      poly(y, 6)4 & -0.8386 & 1.5161 & -0.55 & 0.5802 \\ 
      poly(y, 6)5 & 1.0439 & 1.5157 & 0.69 & 0.4910 \\ 
      poly(y, 6)6 & -6.1972 & 1.5154 & -4.09 & 0.0000 \\ 
       \hline
    \end{tabular}
 }
 \hspace{0.1cm} 
 \subfloat[Model 19 - SAAFB]{
 \centering
    \begin{tabular}{rrrrr}
      \hline
     (Model 19)& Estimate & Std. Error & z value & Pr($>$$|$z$|$) \\ 
      \hline
    (Intercept) & 6.9308 & 0.0262 & 265.00 & 0.0000 \\ 
      poly(x, 3)1 & 58.2700 & 1.5156 & 38.45 & 0.0000 \\ 
      poly(x, 3)2 & -15.0443 & 1.5156 & -9.93 & 0.0000 \\ 
      poly(x, 3)3 & 5.3331 & 1.5156 & 3.52 & 0.0004 \\ 
      poly(y, 6)1 & -5.7890 & 1.5160 & -3.82 & 0.0001 \\ 
      poly(y, 6)2 & -60.7725 & 1.5160 & -40.09 & 0.0000 \\ 
      poly(y, 6)3 & 18.8451 & 1.5157 & 12.43 & 0.0000 \\ 
      poly(y, 6)4 & -0.8444 & 1.5152 & -0.56 & 0.5774 \\ 
      poly(y, 6)5 & 1.0551 & 1.5148 & 0.70 & 0.4861 \\ 
      poly(y, 6)6 & -6.1917 & 1.5145 & -4.09 & 0.0000 \\ 
       \hline
    \end{tabular}
 }
\end{table}

\textcolor{black}{
Models for the factor $edges$ were also obtained for SAAFB and HS. However, as in Sample 1, their results were similar to models with the factor $vertices$ due to the proportionality between $vertices$ and $edges$.
}
\subsubsection{Models for restarted search (Sample 2)}

\begin{figure}[H]
\centering
    \fbox{a.\includegraphics[scale=0.24]{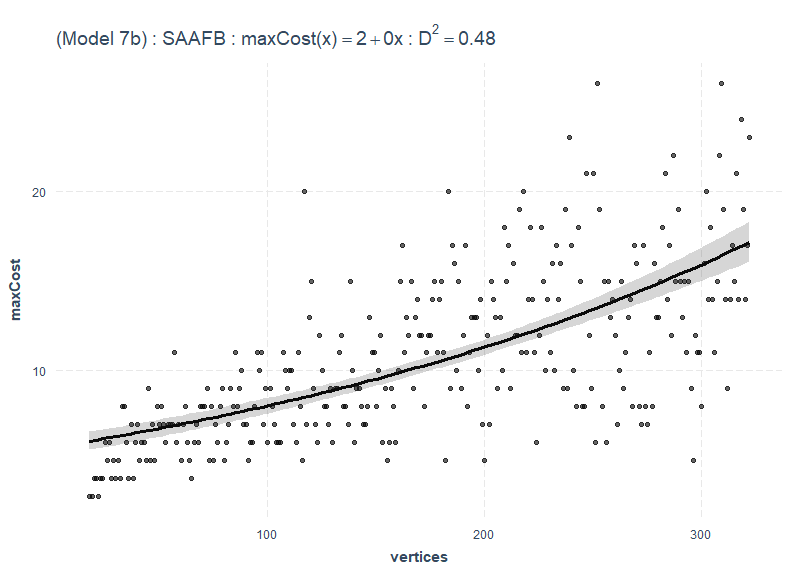}}
    \hspace{5px}
    \fbox{b.\includegraphics[scale=0.24]{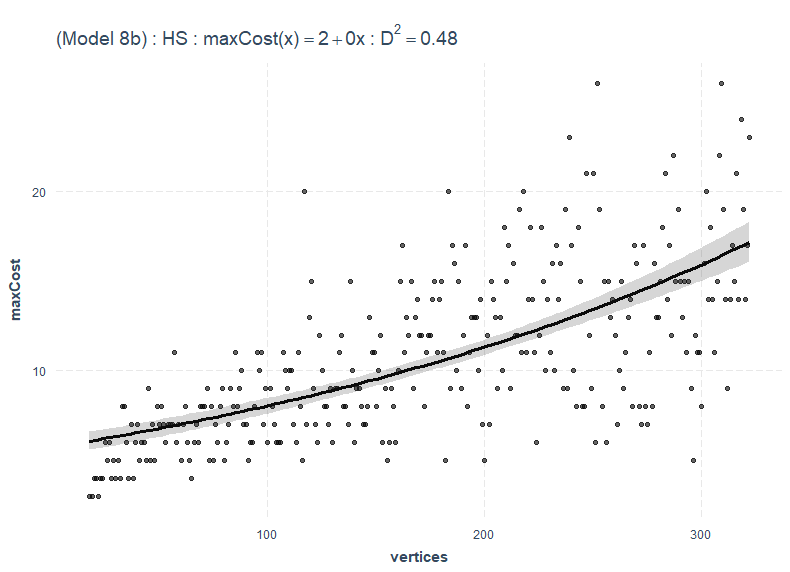}}
    
    \caption{SAAFB and HS $maxCost(vertices)$ for \emph{restarted search} - Sample 2.}
    \label{fig:model7b_8b}
\end{figure}

\textcolor{black}{
For the \emph{restarted search} metric, two models were built, in Sample 2. Models 7b and 8b refer to $maxCost(vertices)$ of SAAFB and HS, as shown in the Figures \ref{fig:model7b_8b}(a) and \ref{fig:model7b_8b}(b). Tables \ref{tab:models7b_8b}(a) and \ref{tab:models7b_8b}(b) display that the models have the same polynomial: $maxCost(x) = 2 + 0x$, where $x$ is $vertices$.
}

\begin{table}[H]
\centering
\tiny
\caption{SAAFB and HS $maxCost(vertices)$ for \emph{restarted search} - Sample 2.}
\label{tab:models7b_8b}
 \subfloat[Model 7b - SAAFB]{
 \centering
 \begin{tabular}{rrrrr}
  \hline
 (Model 7b)& Estimate & Std. Error & z value & Pr($>$$|$z$|$) \\ 
  \hline
    (Intercept) & 1.7439 & 0.0457 & 38.16 & 0.0000 \\ 
      x & 0.0034 & 0.0002 & 15.81 & 0.0000 \\ 
  \hline
\end{tabular}
    
 }
 \hspace{0.1cm} 
 \subfloat[Model 8b - HS]{
 \centering
 \begin{tabular}{rrrrr}
  \hline
 (Model 8b)& Estimate & Std. Error & z value & Pr($>$$|$z$|$) \\ 
  \hline
    (Intercept) & 1.7439 & 0.0457 & 38.16 & 0.0000 \\ 
      x & 0.0034 & 0.0002 & 15.81 & 0.0000 \\ 
   \hline
\end{tabular}
 }
\end{table}

\subsection{Models of Sample 3}

\textcolor{black}{
In Sample 3 (networks between 16 and 80 vertices, but as complete bipartite digraphs without dummies), eight models were elaborated: six referring to the \emph{computational cost} metric and two referring to the \emph{restarted search} metric.
}

\subsubsection{Models for computational cost (Sample 3)}

\textcolor{black}{
Models 11 and 12 refer to $maxCost(vertices)$ of the SAAFB and HS algorithms, as shown in Figures \ref{fig:model11b_12b}(a) and \ref{fig:model11b_12b}(b). Tables \ref{tab:models11_12}(a) and \ref{tab:models11_12}(b) show that the models have the same polynomial: $maxCost(x) = 13 + 31x - 11x^2 + 6x^3$, where $x$ is $vertices$.
}

\begin{table}[H]
\centering
\tiny
\caption{SAAFB and HS $maxCost(vertices)$ for \emph{computational} - Sample 3.}
\label{tab:models11_12}
 \subfloat[Model 11 - SAAFB]{
 \centering
    \begin{tabular}{rrrrr}
      \hline
    (Model 11)& Estimate & Std. Error & t value & Pr($>$$|$t$|$) \\ 
      \hline
    (Intercept) & 13.2331 & 0.0272 & 486.10 & 0.0000 \\ 
      poly(x, 3)1 & 31.5093 & 0.4755 & 66.27 & 0.0000 \\ 
      poly(x, 3)2 & -10.8626 & 0.4755 & -22.85 & 0.0000 \\ 
      poly(x, 3)3 & 5.7247 & 0.4755 & 12.04 & 0.0000 \\ 
       \hline
    \end{tabular}
    
 }
 \hspace{0.1cm} 
 \subfloat[Model 12 - HS]{
 \centering
    \begin{tabular}{rrrrr}
      \hline
    (Model 12)& Estimate & Std. Error & t value & Pr($>$$|$t$|$) \\ 
      \hline
    (Intercept) & 13.2335 & 0.0272 & 486.37 & 0.0000 \\ 
      poly(x, 3)1 & 31.4995 & 0.4752 & 66.29 & 0.0000 \\ 
      poly(x, 3)2 & -10.8533 & 0.4752 & -22.84 & 0.0000 \\ 
      poly(x, 3)3 & 5.7168 & 0.4752 & 12.03 & 0.0000 \\ 
       \hline
    \end{tabular}
 }
\end{table}

\begin{figure}[H]
\centering
    \fbox{a.\includegraphics[scale=0.24]{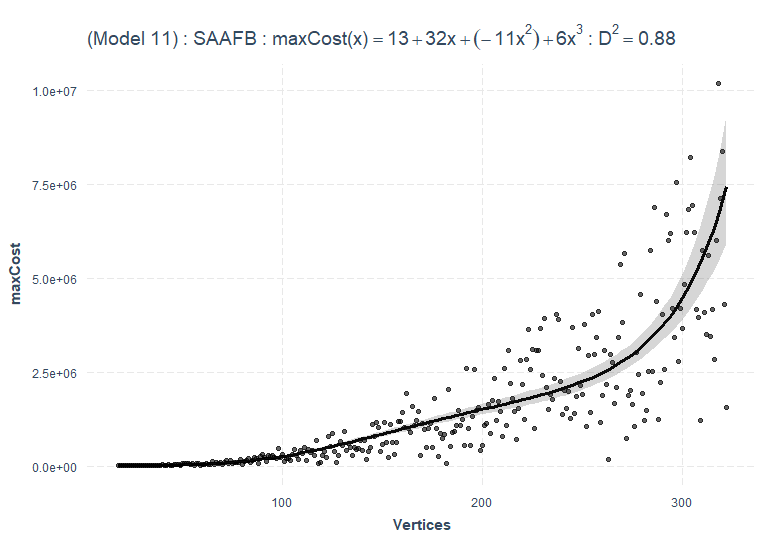}}   
    \hspace{10px}
    \fbox{b.\includegraphics[scale=0.24]{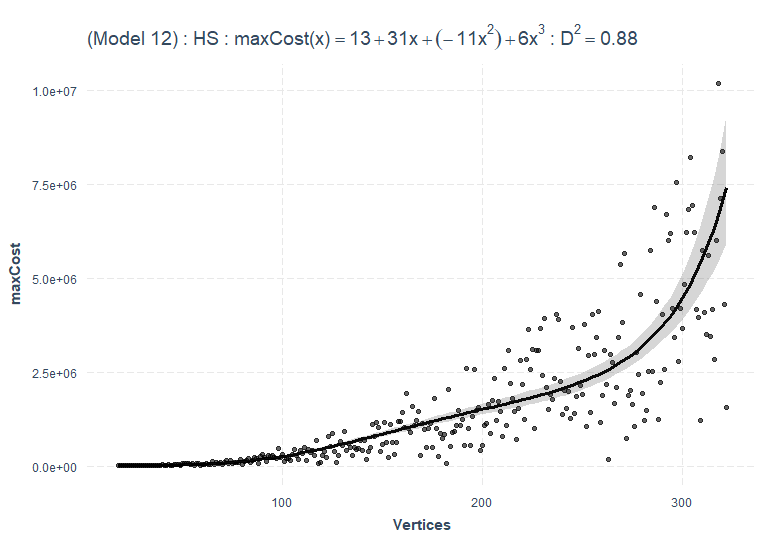}}
    \caption{SAAFB and HS $maxCost(vertices)$ for \emph{computational} - Sample 3.}
    \label{fig:sample3a_one_factor}
\end{figure}

\begin{figure}[H]
\centering
    \fbox{a.\includegraphics[scale=0.24]{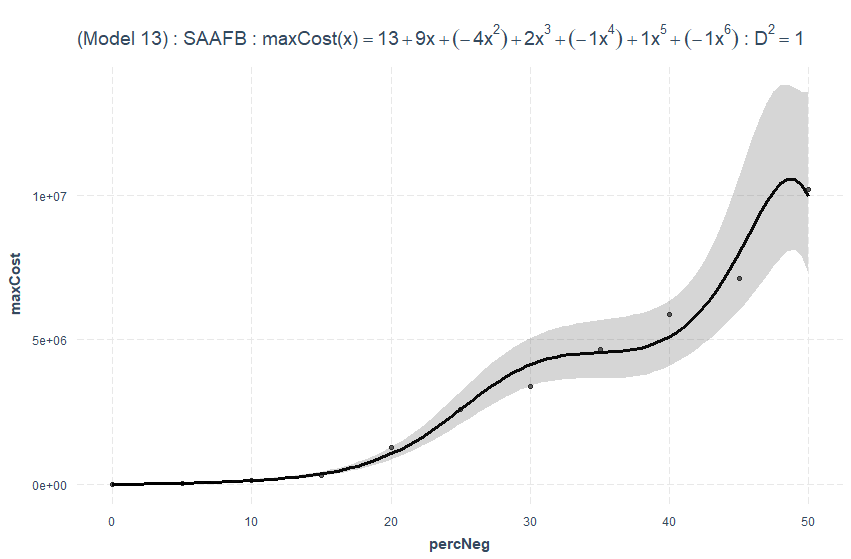}}
    \hspace{10px}
    \fbox{b.\includegraphics[scale=0.24]{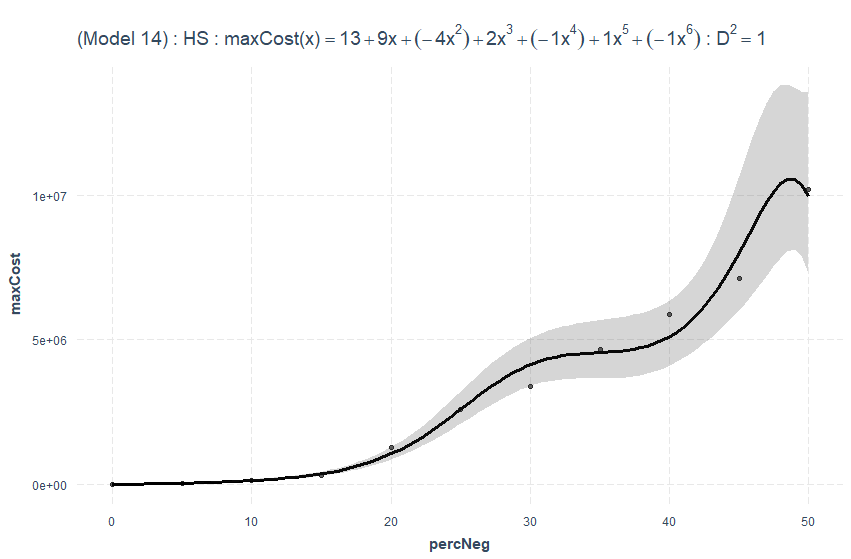}}   
    \caption{SAAFB and HS $maxCost(percNeg)$ for \emph{computational} - Sample 3.}
    \label{fig:sample3b_one_factor}
\end{figure}

\textcolor{black}{
Models 13 and 14 refer to the $maxCost(percNeg)$ of SAAFB and HS, as shown in Figu\-res \ref{fig:sample3b_one_factor}(a) and \ref{fig:sample3b_one_factor}(b). Tables \ref{tab:models13_14}(a) and \ref{tab:models13_14}(b) show that the models have the same polynomial: $maxCost(x) = 13 + 9 - 4^2 + 2x^3 - 1x^4 + 1x^5 - 1x^6$, where $x$ is $percNeg$.
}

\begin{table}[H]
\centering
\tiny
\caption{SAAFB and HS $maxCost(percNeg)$ for \emph{computational} - Sample 3.}
\label{tab:models13_14}
 \subfloat[Model 13 - SAAFB]{
 \centering
    \begin{tabular}{rrrrr}
      \hline
    (Model 13)& Estimate & Std. Error & z value & Pr($>$$|$z$|$) \\ 
      \hline
    (Intercept) & 13.3799 & 0.0341 & 392.41 & 0.0000 \\ 
      poly(x, 6)1 & 8.7585 & 0.1149 & 76.22 & 0.0000 \\ 
      poly(x, 6)2 & -4.0915 & 0.1153 & -35.47 & 0.0000 \\ 
      poly(x, 6)3 & 1.7654 & 0.1147 & 15.40 & 0.0000 \\ 
      poly(x, 6)4 & -0.6835 & 0.1136 & -6.02 & 0.0000 \\ 
      poly(x, 6)5 & 0.7251 & 0.1127 & 6.44 & 0.0000 \\ 
      poly(x, 6)6 & -0.6181 & 0.1121 & -5.51 & 0.0000 \\ 
       \hline
    \end{tabular}
    
 }
 \hspace{0.1cm} 
 \subfloat[Model 14 - HS]{
 \centering
    \begin{tabular}{rrrrr}
      \hline
    (Model 14)& Estimate & Std. Error & z value & Pr($>$$|$z$|$) \\
      \hline
    (Intercept) & 13.3803 & 0.0341 & 392.54 & 0.0000 \\ 
      poly(x, 6)1 & 8.7567 & 0.1149 & 76.23 & 0.0000 \\ 
      poly(x, 6)2 & -4.0899 & 0.1153 & -35.47 & 0.0000 \\ 
      poly(x, 6)3 & 1.7642 & 0.1146 & 15.39 & 0.0000 \\ 
      poly(x, 6)4 & -0.6827 & 0.1135 & -6.01 & 0.0000 \\ 
      poly(x, 6)5 & 0.7246 & 0.1126 & 6.43 & 0.0000 \\ 
      poly(x, 6)6 & -0.6179 & 0.1121 & -5.51 & 0.0000 \\ 
       \hline
    \end{tabular}
 }
\end{table}

\begin{figure}[H]
    \centering
    \fbox{a.\includegraphics[scale=0.24]{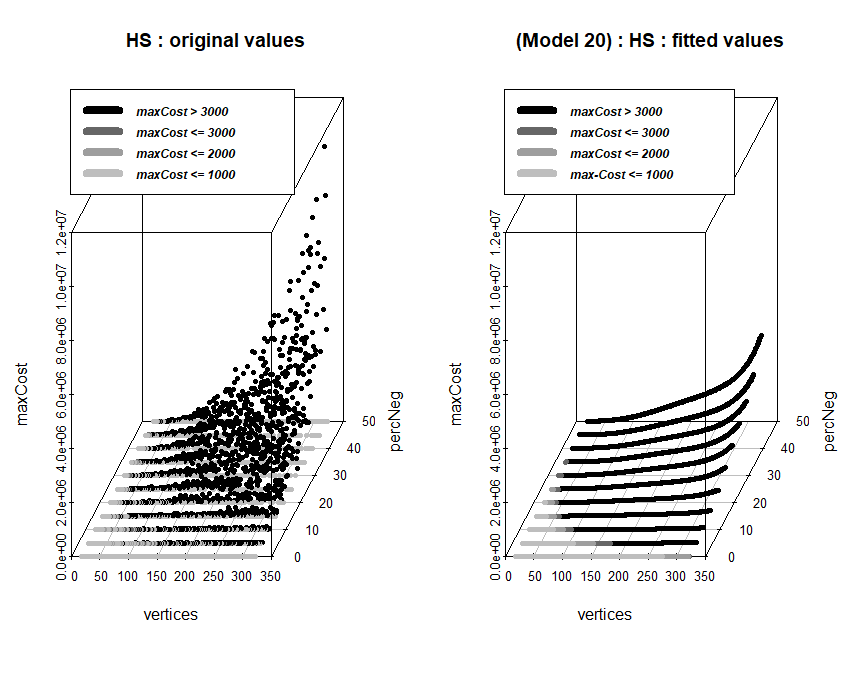}}
    \hspace{5px}
    \fbox{b.\includegraphics[scale=0.24]{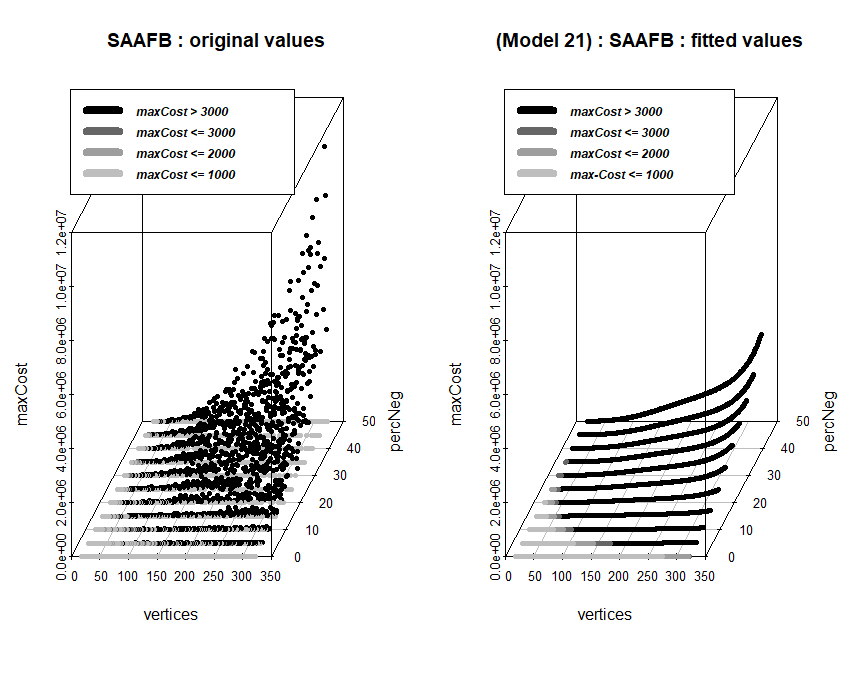}}

    \caption{HS and SAAFB $maxCost(vertices, percNeg)$ for \emph{computational} - Sample 3.}
    \label{fig:sample3c_two_factor}
\end{figure}

\textcolor{black}{
Models 20 and 21 refer to $maxCost(vertices, percNeg)$ of HS and SAAFB, in Sample 3, as shown in Figures \ref{fig:sample3c_two_factor}(a) and \ref{fig:sample3c_two_factor}(b). Tables \ref{tab:models20_21}(a) and \ref{tab:models20_21}(b) demonstrate that the models have the same polynomial: $maxCost(x,y) = 11 + 86x + 120y - 26x^2 - 48y ^2 + 12x^3 + 17y^3 - 7y^4 + 5y^5$, where $x$ is $vertices$ and $y$ is $percNeg$.
}

\begin{table}[H]
\centering
\tiny
\caption{HS and SAAFB $maxCost(vertices, percNeg)$ for \emph{computational} - Sample 3.}
\label{tab:models20_21}
 \subfloat[Model 20 - HS]{
 \centering
    \begin{tabular}{rrrrr}
      \hline
      (Model 20)& Estimate & Std. Error & z value & Pr($>$$|$z$|$) \\ 
      \hline
(Intercept) & 10.5202 & 0.0386 & 272.89 & 0.0000 \\ 
  poly(x, 3)1 & 85.7117 & 2.2336 & 38.37 & 0.0000 \\ 
  poly(x, 3)2 & -25.1565 & 2.2337 & -11.26 & 0.0000 \\ 
  poly(x, 3)3 & 11.9616 & 2.2336 & 5.36 & 0.0000 \\ 
  poly(y, 5)1 & 119.7732 & 2.2337 & 53.62 & 0.0000 \\ 
  poly(y, 5)2 & -47.7613 & 2.2338 & -21.38 & 0.0000 \\ 
  poly(y, 5)3 & 17.1911 & 2.2336 & 7.70 & 0.0000 \\ 
  poly(y, 5)4 & -7.4775 & 2.2332 & -3.35 & 0.0008 \\ 
  poly(y, 5)5 & 5.3063 & 2.2328 & 2.38 & 0.0175 \\ 

      \hline
    \end{tabular}
    
 }
 \hspace{0.1cm} 
 \subfloat[Model 21 - SAAFB]{
 \centering
    \begin{tabular}{rrrrr}
      \hline
        (Model 21)& Estimate & Std. Error & z value & Pr($>$$|$z$|$) \\ 
      \hline
    (Intercept) & 10.5172 & 0.0385 & 272.84 & 0.0000 \\ 
      poly(x, 3)1 & 85.8663 & 2.2333 & 38.45 & 0.0000 \\ 
      poly(x, 3)2 & -25.3112 & 2.2335 & -11.33 & 0.0000 \\ 
      poly(x, 3)3 & 12.0930 & 2.2334 & 5.41 & 0.0000 \\ 
      poly(y, 5)1 & 119.9224 & 2.2335 & 53.69 & 0.0000 \\ 
      poly(y, 5)2 & -47.8990 & 2.2336 & -21.44 & 0.0000 \\ 
      poly(y, 5)3 & 17.2933 & 2.2333 & 7.74 & 0.0000 \\ 
      poly(y, 5)4 & -7.5425 & 2.2329 & -3.38 & 0.0007 \\ 
      poly(y, 5)5 & 5.3410 & 2.2326 & 2.39 & 0.0167 \\ 
       \hline
    \end{tabular}
 }
\end{table}

\textcolor{black}{
Models for the $edges$ factor were also tested. However, their results showed low explanatory power with $D^2$.}

\subsubsection{Models for restarted search (Sample 3)}

\begin{figure}[H]
\centering
    \fbox{a.\includegraphics[scale=0.24]{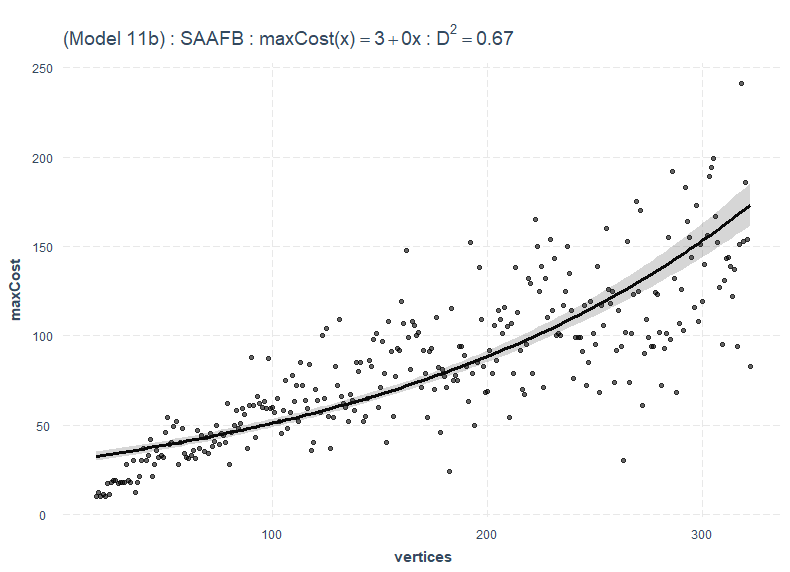}}
    \hspace{5px}
    \fbox{b.\includegraphics[scale=0.24]{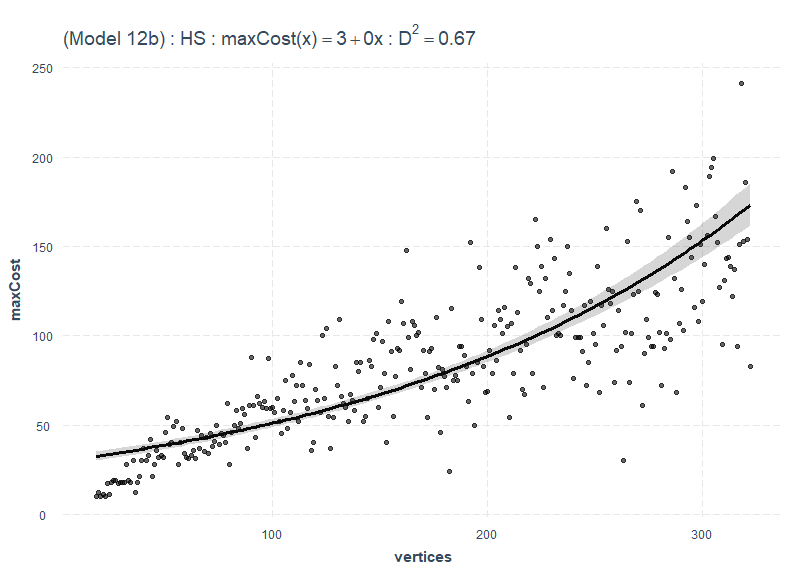}}
    
    \caption{SAAFB and HS $maxCost(vertices)$ for \emph{restarted search} - Sample 3.}
    \label{fig:model11b_12b}
\end{figure}

\textcolor{black}{
For the \emph{restarted search} metric, in Sample 3, two more models were built. Models 11b and 12b refer to $maxCost(vertices)$ of SAAFB and HS, as shown in Figures \ref{fig:model11b_12b}(a) and \ref{fig:model11b_12b}(b). Tables \ref{tab:models11b_12b}(a) and \ref{tab:models11b_12b}(b) show that the models have the same polynomial: $maxCost(x) = 3 + 0x$, where $x$ is $vertices$.
}

\begin{table}[H]
\centering
\tiny
\caption{SAAFB and HS $maxCost(vertices)$ for \emph{restarted search} - Sample 3.}
\label{tab:models11b_12b}
 \subfloat[Model 11b - SAAFB]{
 \centering
 \begin{tabular}{rrrrr}
  \hline
 (Model 11b)& Estimate & Std. Error & z value & Pr($>$$|$z$|$) \\ 
  \hline
    (Intercept) & 3.3821 & 0.0404 & 83.65 & 0.0000 \\ 
      x & 0.0055 & 0.0002 & 26.74 & 0.0000 \\ 
  \hline
\end{tabular}
    
 }
 \hspace{0.1cm} 
 \subfloat[Model 12b - HS]{
 \centering
 \begin{tabular}{rrrrr}
  \hline
 (Model 12b)& Estimate & Std. Error & z value & Pr($>$$|$z$|$) \\ 
  \hline
    (Intercept) & 3.3821 & 0.0404 & 83.65 & 0.0000 \\ 
      x & 0.0055 & 0.0002 & 26.74 & 0.0000 \\
   \hline
\end{tabular}
 }
\end{table}

\section{Analysis and discussion}\label{chpt:analysis_discussion}

\textcolor{black}{
The results of models are summarized by sample in Tables  \ref{tab:highest_degree_sample1}, \ref{tab:highest_degree_sample2}, and \ref{tab:highest_degree_sample3}. These results are organized by metric, factors of the $maxCost()$ function ($vertices$; $percNeg$; and $vertices$ with $percNeg$), and algorithms (RSFB, SAAFB and HS). The results are reduced to terms with the highest degree of the respective polynomials, highlighting only the order of magnitude in each case. It is important to remember that the \emph{computational cost} metric refers to the total cost of each algorithm and the \emph{restarted search} metric refers to the number of times the search is restarted in the spanning tree by each algorithm (open questions for the three algorithms).
}

\subsection{Discussion about Sample  1}
\textcolor{black}{
Table \ref{tab:highest_degree_sample1} summarize all models of Sample 1, where the first three rows show the models for the \emph{computational cost} metric (as \emph{comp. cost}) and the last row shows the models for the \emph{restarted search} metric (as \emph{rest. search}). So, in first row, for \emph{computational cost} with $maxCost(vertices)$ the performance of RSFB was $vertices^3$ and performance of SAAFB and HS were $vertices^2$. In the second row for \emph{computational cost} with $maxCost(percNeg)$ all algorithms showed the same performance, i.e., $percNeg^4$. In the third row for the \emph{computational cost} with $maxCost(vertices, percNeg)$ all results were $vertices^2+percNeg^6$. In the last row, for the \emph{restarted search} with $maxCost(vertices)$ all algorithms showed the same results, i.e., $vertices^1$.\\
}

\textcolor{black}{
It should be noted that the KS test, described in Table \ref{tab:KS_test}, pointed out that, in Sample 1 for the  \emph{computatinal cost}: RSFB $\not\sim$ SAAFB; RSFB $\not \sim$ SAAFB; and HS $\sim$ SAAFB. Such test reinforces the results of Table \ref{tab:highest_degree_sample1}. In other words, the polynomial degrees when the metric is \emph{computational cost} and the $maxCost()$ functions of RSFB diverge from the degrees of the respective functions of SAAFB and HS. However, it is noticed that between SAAFB and HS, comparing by function $maxCost()$, the polynomial degrees are always the same.
}

\begin{table}[H]
\centering
\caption{Highest degree of polynomials by metric - Sample 1.}
\label{tab:highest_degree_sample1}
\begin{tabular}{ccccc}
  \hline
 Metric & $maxCost$ Factors & RSFB degree  & SAAFB degree & HS degree\\ 
  \hline
  \emph{comp. cost}&\emph{vertices} & $vertices^3$ & $vertices^2$ & $vertices^2$ \\ 
  \emph{comp. cost}&\emph{percNeg} & $percNeg^4$ & $percNeg^4$ & $percNeg^4$ \\ 
  \emph{comp. cost}&\emph{vertices, percNeg} & $vert.^2+percNeg^6 $ & $vert.^2+percNeg^6$ & $vert.^2+percNeg^6$ \\
  \emph{rest. search}&\emph{vertices} & $vertices^1$ & $vertices^1$ & $vertices^1$ \\ 
   \hline
\end{tabular}
\end{table}

\subsection{Discussion about Sample 2}
\textcolor{black}{
Similar to the previous section, the results of the models in Sample 2 are summarized\- in Table \ref{tab:highest_degree_sample2}. In the first row, for \emph{computational cost} with $maxCost(vertices)$, the results of SAAFB and HS were $vertices^3$. In the second row for \emph{computational cost} with $maxCost(percNeg)$ the results of SAAFB and HS were $percNeg^6$. In the third row for the \emph{computational cost} with $maxCost(vertices, percNeg)$ the results were $vertices^2+percNeg^6$. In the last row, for the \emph{restarted search} with $maxCost(vertices)$, the results of SAAFB and HS were  $vertices^1$, as well as in Sample 1.\\
}

\textcolor{black}{
With respect to the KS test, according to Table \ref{tab:KS_test} (in Sample 2) for the \emph{computational cost}: SAAFB $\sim$ HS. This result is in harmony with the results of Table \ref{tab:highest_degree_sample2}. In other words, the maximum degrees of the polynomials are the same in all $maxCost()$ functions between SAAFB and HS.
}

\begin{table}[H]
\centering
\caption{Highest degree of polynomials by metric - Sample 2.}
\label{tab:highest_degree_sample2}
\begin{tabular}{cccc}
  \hline
 Metric& $maxCost$ Factors &  SAAFB degree & HS degree\\ 
  \hline
  \emph{computational cost}&\emph{vertices} & $vertices^3$ & $vertices^3$ \\ 
  \emph{computational cost}&\emph{percNeg} & $percNeg^6$ & $percNeg^6$ \\ 
  \emph{computational cost}& $vertices,percNeg$ & $ vertices^3 + percNeg^6$ & $ vertices^3 + percNeg^6$\\
  \emph{restarted search}&\emph{vertices} & $vertices^1$ & $vertices^1$ \\ 
   \hline
\end{tabular}
\end{table}

\subsection{Discussion about Sample  3}
\textcolor{black}{
In Sample  3, for \emph{computational cost} with $maxCost(vertices)$, through Table \ref{tab:highest_degree_sample3}, it is noted that the polynomials of SAAFB and HS do not exceed degree 3. In the third row, for the \emph{computational cost} with $maxCost(vertices, percNeg)$, the results were $vertices^2+percNeg^5$. The other results were the same as in Sample 2.
}

\begin{table}[ht]
\centering
\caption{Highest degree of polynomials by metric - Sample 3.}
\label{tab:highest_degree_sample3}
\begin{tabular}{cccc}
  \hline
 Metric& $maxCost$ Factors & SAAFB degree & HS degree\\ 
  \hline
  \emph{computational cost} & \emph{vertices} & $vertices^3$ & $vertices^3$\\ 
  \emph{computational cost} & \emph{percNeg} & $percNeg^6$ & $percNeg^6$ \\ 
  \emph{computational cost} & \emph{vertices, percNeg} & $vertices^3+percNeg^5$ & $vertices^3+percNeg^5$ \\
  \emph{computational cost} & \emph{vertices}  & $vertices^1$ & $vertices^1$\\
   \hline
\end{tabular}
\end{table}\unskip
\section{Conclusions}\label{chpt:future}
\textcolor{black}{
In this article, the most recent algorithms (RSFB, SAAFB, and HS) for the problem described in Section 2 were presented and submitted to a factorial experiment. The experiment structure employed a function called $maxCost$, whose parameters were network factors such as $vertices$ and $percNeg$ (percentage of negative activities). The results were expressed through two metrics: 1) \emph{computational\- cost} and 2) \emph{restarted search}. While the first metric referred to the total cost for each of the algorithms, the second metric referred to the number of search restarts in the \emph{spanning tree} for each algorithm.\\
}

\textcolor{black}{
The results were obtained using three samples from different network configurations. In the first sample, graphs with 16 to 80 vertices were used, randomized in terms of vertices, layers, and percentage of negative activities, among other parameters. The second sample used graphs with 16 to 320 vertices and randomization similar to the first sample. Finally, the third sample referred to a set of convenience graphs to stress the algorithms with a high number of edges. In this case, complete bipartite digraphs were used (disregarding $dummies$) with vertices between 16 and 320.
}

\subsection{Main Conclusions}
\subsubsection{SAAFB and HS outperform RSFB for the first metric}
\textcolor{black}{
As indicated, the RSFB algorithm was used only in the networks of Sample 1. In this case, it is worth noting that the highest degree of the polynomial identified for the metric \emph{computational cost} with $maxCost(vertices)$ was $vertices^3$, while the algorithms SAAFB and HS were expressed in polynomials of degree two, i.e., $vertices^2$. The empirical results show that by using $maxCost(vertices)$ as a proxy for $O(vertices)$, we can estimate that RSFB has a time cost of $O(vertices^3)$, while SAAFB and HS presented a time cost of $O(vertices^2)$, with the networks in Sample 1. In addition, the results of the KS test, according to Table \ref{tab:KS_test}, showed a statistically significant difference between the distributions of RSFB and SAAFB; and RSFB and HS.
}

\subsubsection{SAAFB and HS have the same performance for both metrics}
\textcolor{black}{
In the case of SAAFB and HS, the results of the KS test (in the three samples and both metrics) did not present any statistically significant difference, as shown by the factor comparisons of the $maxCost$ function (Table \ref{tab:KS_test}). Furthermore, they presented equal values for the maximum degree of polynomials for all factors and metrics analyzed. So we can conclude, within the limits of statistical precision, that both  SAAFB and HS present the same order of cost performance.
}

\subsubsection{RSFB, SAAFB and HS have similar performance for the second metric}
\textcolor{black}{
The results of the \emph{restarted search} metric in all samples by function $maxCost$ of the three algorithms were always the same. In this case, the highest degree of the polynomials of the elaborated models was always $vertices^1$. In other words, RSFB, SAAFB, and HS presented a time cost of $O(vertices^1)$. It is an important finding, as it refers to one of the open questions in the three algorithms.
}
\subsubsection{Performance most influencial factor}
\textcolor{black}{
The models that include the $percNeg$ factor presented polynomials with higher degrees than those that do not have such a factor. However, the $percNeg$ factor considers a range with values from 0\% to 100\%. In this case, it was possible to verify that the influence of $percNeg$ on the scheduling cost increases until the value is around 50\%, after which the influence decreases. On the other hand, the $vertices$ factor influences the cost of scheduling without a defined limit. Overall, this indicates that as long as the $vertices$ factor can grow, the scheduling cost will also grow without limit. Therefore, the factor $vertices$ exclusively is more appropriate to express the order of cost growth of the algorithms in the three samples with the two metrics. In this sense, where $n$ is $vertices$, the cost of scheduling was: a) $O(n^3)$ for RSFB, with the \emph{computational cost} metric in Sample 1; b) $O(n^2)$, with the \emph{computational cost} metric for SAAFB and HS, in Sample 1; c) $O(n^3)$, with the \emph{computational cost} metric for SAAFB and HS, in Samples 2 and 3; d) and $O(n)$, with the \emph{restarted search} metric for RSFB, SAAFB and HS, in the three samples.
}

\subsection{Future Research}
\textcolor{black}{
Among the possibilities for future research are: 1) consider samples with larger graphs than those treated in this experiment (above 320 vertices) and graphs with redundant edges by transiti\-vity. Such cases can subject the algorithms to higher stress levels; 2) implement the RSFB algorithm in some language that supports double recursive stacking (characteristic of the algorithm) in networks with 320 (or more) vertices. Then perform a new experiment. With networks in the order of 320 (or more) vertices, it is conjectured that the order of cost of RSFB is greater than $O(n^3)$, which was verified in this experiment for the algorithms SAAFB and HS.\\
}

\section{Acknowledgements}
\textcolor{blue}{
}
\textcolor{black}{
We are grateful for the support offered by the Graduate Program in Informatics of the Mathematics Institute of the Federal University of Rio de Janeiro (PPGI-IM-UFRJ) in Brazil, the Universal Project of the National Council for Scientific and Technological Development in Brazil (CNPq), and the Institute of Computing of the Federal University of Amazonas (IComp-UFAM) in Brazil.
}




\begin{thebibliography}{10}



\bibitem[Battersby(1964)]{B:64}Battersby, A. (1964). Network analysis for planning and scheduling. Macmillan; St Martin's Press, New York.
\bibitem[Blazewicz \textit{et al.}(1983)]{B:L:R83}Blazewicz J., Lenstra J., \& Rinnooy-Kan A. (1983). Scheduling subject to resource constraints: classification and complexity. Discrete applied mathematics, 5(1), 11-24.
\bibitem[Cavalcante \textit{et al.}(1998)]{CDSSWW:01}Cavalcante, C. C., De Souza, C. C., Savelsbergh, M. W., Wang, Y., \& Wolsey, L. A. (2001). Scheduling projects with labor constraints. Discrete Applied Mathematics, 112(1-3), 27-52.
\bibitem[Chang \& Edmonds(1985)]{CE:85}Chang, G. J., \& Edmonds J. (1985). The poset scheduling problem. Order, 2(2), 113-118.
\bibitem[Dantzig(1963)]{DANTZIG:63}Dantzig, G. B. (1963). Linear Programming and Extensions. Princeton University Press, New Jersey.
\bibitem[Demeulemeester \textit{et al.}(1996)]{DHV:96}Demeulemeester, E. L., Herroelen, S. W., \& Van Dommelen, P. (1996). An Optimal Recursive Search Procedure for the Deterministic Unconstrained max-npv Project Scheduling Problem. Katholieke Universiteit Leuven.
\bibitem[Demeulemeester \textit{et al.}(1997)]{D:H97}Demeulemeester, E. L., Herroelen, S. W., \& De Reyck B. (1997).  A classification scheme for project scheduling problems.  Techinical Report 2, Katholieke Universiteit Leuven.
\bibitem[Demeulemeester \& Herroelen (2002)]{DH:02}Demeulemeester, E. L., \& Herroelen, S. W. (2002).  Project Scheduling: A research handbook. Kluwer’s International Series, New York.
\bibitem[Elmaghraby \& Herroelen(1990)]{EH:90}Elmaghraby, S. E., \& Herroelen, W. S. (1990). The scheduling of activities to maximize the net present value of projects. European Journal of Operational Research, 49(1), 35-49.
\bibitem[Erdõs \& Rényi(1960)]{ER:60}Erdős, P., \& Rényi, A. (1960). On the evolution of random graphs. Publ. Math. Inst. Hung. Acad. Sci, 5(1), 17-60.
\bibitem[Fávero \& Belfiore (2019)]{FP:19}Fávero, L., \& Belfiore, P. (2019). Data science for business and decision making. Academic Press.
\bibitem[Graham \textit{et al.}(1979)]{GLLR:79}Graham, R. L., Lawler, E. L., Lenstra, J. K., \& Kan, A. R. (1979). Optimization and approximation in deterministic sequencing and scheduling: a survey. In Annals of discrete mathematics (Vol. 5, pp. 287-326). Elsevier.
\bibitem[Grinold(1972)]{G:72}Grinold, R. C. (1972). The payment scheduling problem. Naval Research Logistics Quarterly, 19(1), 123-136.
\bibitem[Gröflin \textit{et al.}(1982)]{GLP:82}Gröflin, H., Liebling, T., \& Prodon, A. (1982). Optimal Subtrees and Extensions. Annals of Discrete Math. vol.16, 121-127.
\bibitem[Herroelen \textit{et al.}(1997)]{H:D:VD97}Herroelen, W. S., Van Dommelen, P., \& Demeulemeester, E. L. (1997). Project network models with discounted cash flows a guided tour through recent developments. European Journal of Operational Research, 100(1), 97-121.
\bibitem[Herroelen \& Gallens(1993)]{HG:93}Herroelen, W. S., \& Gallens, E. (1993). Computational experience with an optimal procedure for the scheduling of activities to maximize the net present value of projects. European Journal of Operational Research, 65(2), 274-277.


\bibitem[Nelder \& Wedderburn (1972)]{N:W72}Nelder, J. A., \& Wedderburn, R. W. (1972). Generalized linear models. Journal of the Royal Statistical Society: Series A (General), 135(3), 370-384.
\bibitem[Juang(1994)]{JUANG:94}Juang, S. H. (1994). Optimal solution of job-shop scheduling problems – a new network flow approach. Phd thesis, The University of Iowa, USA.
\bibitem[Kazaz \& Sepil(1996)]{KS:96}Kazaz, B. \& Sepil, C. (1996). Project Scheduling with Discounted Cash Flows and Progress Payments. Journal of the Operational Research Society, 47(10), 1262–1272.
\bibitem[Margot \textit{et al.}(1990)]{MPL:90}Margot, F., Prodon, A., \& Liebling, T. M. (1990). The poset scheduling problem. Methods of Operations Research, 62, 221-230.
\bibitem[Roundy \textit{et al.}(1991)]{RMHTG:91}Roundy, R.,  Maxwell, W.; Herer, Y., Tayur, S., \& Getzler, A. (1991). A price-directed approach to real-time scheduling of productionoperation. IIE Transac., 23, 149–160.
\bibitem[Russell(1970)]{R:70}Russell, A. (1970). Cash flows in networks. Management Science, 16, 357–373.
\bibitem[Schwindt \& Zimmermann (2001)]{SZ:01}
Schwindt, C., \& Zimmermann, J.  (2001). A steepest ascent approach to maximizing the net present value of projects. Mathematical Methods of Operations Research 53, 435-450.
\bibitem [Sankaran \textit{et al.}(1999)]{SBJ:99}Sankaran, J. K., Bricker, D. L., \& Juang, S. H. (1999). A strong fractional cutting-plane algorithm for resource-constrained project scheduling. International Journal of Industrial Engineering, 6, 99-111.
\bibitem[Smith-Daniels(1986)]{SD:86}Smith-Daniels, D. E. (1986). Summary measures for predicting the net present value of a project. In College of St. Thomas St. Paul, Minnesota Working Paper.
\bibitem[Vanhoucke(2006)]{V:06}Vanhoucke, M. (2006). An efficient hybrid search algorithm for various optimization problems. In European Conference on Evolutionary Computation in Combinatorial Optimization (pp. 272-283). Springer, Berlin, Heidelberg.
\bibitem[Vanhoucke \textit{et al.}(1999)]{VDH:99}Vanhoucke, M., Demeulemeester, E., \& Herroelen, W. (1999). On maximizing the net present value of a project under resource constraints. K.U.Leuven - Departement toegepaste economische wetenschappen. 1–24 p. Avaliable in: https://lirias.kuleuven.be/retrieve/60236.
\bibitem[Vanhoucke \textit{et al.}(2000)]{VDH:00}Vanhoucke, M., Demeulemeester, E., \& Herroelen, W. (2000). A validation of procedures for maximizing the net present value of a project. DTEW Research Report 0030.
\bibitem[Vanhoucke \textit{et al.}(2001)]{V:D:H01}Vanhoucke, M., Demeulemeester, E., \& Herroelen, W. (2001). On maximizing the net present value of a project under renewable resource constraints. Management Science. 47, 1113-1121.


\end{thebibliography}



\newpage
\appendix
\section{Experiment algorithms in $backward$}\label{chpt:appA}
\subsection{Recursive Search Forward-Backward \emph{(RSFB)}}

\begin{algorithm}[H]
    \label{algo:Step_3_b}
    \SetAlgoLined
    \normalsize
    \small
     \textbf{procedure} \textit{Step\_3}() 
         $\left\{\textit{CA is a global structure}\right\}$\\
    \Indp
         \textcolor{blue}{$total\_Step\_3  = total\_Step\_3$} + 1\\
         $CA \leftarrow \varnothing$ \\  
         \underline{$SA', DC' \leftarrow$ \emph{Recursion}(n)} \\
        Report the optimal solution $DC'$\\
    \Indm

    \caption{\emph{Step\_3 - Backward.}}
\end{algorithm}

\begin{algorithm}[H]
    \label{algo:RecursionRS_b}
     \normalsize
     \small
     \textbf{function} \emph{Recursion}($newnode$)  \quad $\left\{\textit{CA, CT are global structures}\right\}$ \ \quad \quad \quad \quad  
     \\
    \SetAlgoNoLine
     \Indp
         \textcolor{blue}{$total\_Recursion  = total\_Recursion$} + 1\\
         $SA \leftarrow \left\{newnode\right\}; DC \leftarrow DC_{newnode}; CA \leftarrow CA + newnode$ \\ 
    
         \For{\Each$(i | i \notin CA \ \textbf{\emph{and}} \ i \ \underline{\emph{precedes} \ newnode \in CT})$ }
         {
            $SA', DC' \leftarrow Recursion(i)$   \\
            \eIf{$DC' \geq 0$}{$SA  \leftarrow SA + SA';DC  \leftarrow DC + DC'$}{$\underline{CT \leftarrow CT - (i, newnode)}$ \\
            $\emph{Compute} v_{l*k*} = min \left\{\underline{s_k - f_l} \right\}; \underline{CT \leftarrow CT + (l*,k*)}$ \\
            $\quad \quad \quad \quad \quad \quad \quad \quad ^{\underline{(l*,k*) \in G}}$\\
            $\quad \quad \quad \quad \quad \quad \quad \quad ^{\underline{k* \notin SA}}$\\
            $\quad \quad \quad \quad \quad \quad \quad \quad ^{\underline{l* \in SA}}$\\ 
            
            $\forall \ j \in SA' : \underline{f_j \leftarrow f_j - v_{l*k*}}$ \\
            \emph{Step\_3}()
            }
         }
        \For{\Each$(i | i \notin CA \ \textbf{\emph{and}} \ i \ \underline{\emph{succeeds} \ newnode \in CT})$ }{
         $SA', DC' \leftarrow \emph{Recursion}(i)$  \\ 
         $SA \leftarrow SA + SA';DC \leftarrow DC + DC'$ \\}
    
    \Return($SA, DC$)\\
    \Indm
    
    \caption{\emph{Recursion - Backward.}}
\end{algorithm}

\subsection{Steepest Ascent Approach Forward-Backward \emph{(SAAFB)}}

\SetKwComment{Comment}{/* }{ */}
\begin{algorithm}[H]
    \normalsize
    \small
	 $\textbf{function} \ \emph{SAD}()$  \quad $\left\{ST(V_{st}, E_{st})\textit{ is a global structure}\right\}$\\
     \SetAlgoLined
     \Indp
        \SetAlgoNoLine
         $Z \leftarrow \varnothing$; $V \leftarrow V_{st}$\\ 
         $\forall \ i \in V \ \mathbf{do} \ C(i) \leftarrow  {i}$; $\phi_{i} \leftarrow -\alpha \ c_{i} \ e^{-\alpha^{(S_{i} + d_{i})}}$\\
        \While{V $\neq $ $\left\{ 1 \right\}$}{
                \eIf{$(V \emph{ has a \underline{node sink} $i$ $\neq 1$}) \And (\emph{at most \underline{one predecessor}}\ j)$}{
                \textcolor{blue}{$iteration\_SAD  = iteration\_SAD$} + 1\\
                $C(j) \leftarrow  C(j) + C(i)$;
                 $\phi_{j} \leftarrow \phi_{j} + \phi_{i}$;
                 $V \leftarrow V - i$ \\
                }{
                 \If{$(V \emph{ has a \underline{node source} $j$ $\neq 1$}) \And (\emph{only \underline{one successor}}\ i)$}{
                    \textcolor{blue}{$iteration\_SAD  = iteration\_SAD$} + 1\\
                     $\mathbf{if} \ \underline{\phi_{j} \leq 0} \ \mathbf{then} \ Z \leftarrow Z + C(j)$ \\
                    $\mathbf{else}$ \ $\phi_{i} \leftarrow \phi_{i} + \phi_{j}$; \ $C(i) \leftarrow C(i) + C(j)$; \ 
                    $V \leftarrow V - j$\\
                    }
                }
           }
        $\Return(Z)$\\
    \Indp
     \caption{\emph{Steepest Ascent Direction} (SAD) - \emph{Backward.}}
     \label{algo:SAD_b}
\end{algorithm}

\begin{algorithm}[H]
    \label{algo:VA_b}
     \SetAlgoNoLine
     \normalsize
     \small
     \textbf{function} \emph{VA}(\emph{S, Z}) \quad $\left\{\textit{ST, G are global structures}\right\}$\\
     \Indp 
         $\forall \ (i,j) \in ST \ | \ (j \notin C(i)) \And (i \notin C(j)) : ST \leftarrow ST - (i, j)$ \\
         \While{$Z \neq \varnothing$}{
          $\emph{Compute} V_{l*k*} = min \left\{\underline{S_k - f_l} \right\}$\\
           $\quad \quad \quad \quad \quad \quad \quad \quad \underline{^{(l*,k*) \in G}}$\\
           $\quad \quad \quad \quad \quad \quad \quad \quad \underline{^{k* \notin Z}}$\\
           $\quad \quad \quad \quad \quad \quad \quad \quad \underline{^{l* \in Z}}$\\
           
           Take the set $C(j)$ $\in Z$ where $k*$ is contained\\
    
            $\forall \ i \in C(j) : S_i \leftarrow \underline{S_i - V_{l*k*}}$\\
            $Z \leftarrow Z - C(j)$; \
            $ST \leftarrow \underline{ST + (l*,k*)}$\\
           }
        $\Return(\emph{S})$\\
    \Indm
     \caption{\emph{Vertex Ascent} (VA) - \emph{Backward.}}
\end{algorithm}

\begin{algorithm}[H]
     \SetAlgoNoLine
     \normalsize
     \small
     \textbf{procedure} \emph{SAP}() \quad $\left\{\textit{ST is a global structure}\right\}$\\
     \Indp
         $S, ST \leftarrow $ \underline{Determine the Late Schedule ($S$)} as a vector and a corresponding initial Spanning Tree ($ST$) through the original graph $G$. \\
         \textcolor{blue}{$total\_SAD  = 0$}\\
         $Z \leftarrow$ \emph{SAD}()\\
         \While{$Z \neq \varnothing$}{
           \textcolor{blue}{$total\_SAD  = total\_SAD$} + 1\\
           $S \leftarrow$ \emph{VA(S, Z)}\\
           $Z \leftarrow$ \emph{SAD}()\\
           }
           Report the optimal solution $S$\\
     \Indm  
     \caption{\emph{Steepest Ascent Procedure} (SAP) - \emph{Backward.}}
     \label{algo:SAP_b}
\end{algorithm}

\subsection{Hybrid Search \emph{(HS)}}

\begin{algorithm}[H]
     \SetAlgoNoLine
     \normalsize
     \small
     \textbf{function} \emph{Recursion}($newnode$) \quad $\left\{\textit{CA, ST, SS are global structures}\right\}$\\
     \Indp
         \textcolor{blue}{$total\_Recursion  = total\_Recursion$} + 1\\
         $SA \leftarrow \left\{newnode\right\}; DC \leftarrow DC_{newnode}; \ CA \leftarrow CA + newnode$\\
         \For{\Each$(i | i \notin CA \ \textbf{\emph{and}} \ \underline{i \ \emph{precedes} \ newnode \in ST})$}
         {
            $SA', DC' \leftarrow \emph{Recursion}(i)$\\
            \eIf{\underline{$DC' < 0$}}{$SA \leftarrow SA + SA';DC \leftarrow DC + DC'$}{$ST \leftarrow \underline{ST - (i, newnode)}; \ SS \leftarrow SS + SA'$}
         }
         
        \For{\Each$(i | i \notin CA \textbf{\emph{and}} \underline{i \emph{succeeds} newnode \in ST})$}
         {
             $SA', DC' \leftarrow \emph{Recursion}(i)$\\ 
             $SA \leftarrow SA + SA';DC \leftarrow DC + DC'$\\
         }
         \Return(\emph{SA, DC})\\
    \Indm
    \caption{\emph{Recursion} de HS - $Backward$.}
    \label{algo:RecursionHS_b}
\end{algorithm}

\begin{algorithm}[H]
     \SetAlgoNoLine
     \normalsize
     \small
     \textbf{procedure} \emph{Shift\_activities}() \quad
     $\left\{\textit{SS, ST, and G are global structures}\right\}$\\
     \Indp
         $Z \leftarrow \varnothing; \ \forall \ i \in SA \ | \ SA \in SS: Z \leftarrow Z + i$ \\
         
         \While{$Z \neq \varnothing$}{
    
           $\emph{Compute} v_{l*k*} = min \left\{\underline{s_k - f_l} \right\}$\\
           $\quad \quad \quad \quad \quad \quad \quad \quad \underline{^{(l*,k*) \in G};}$\\
           $\quad \quad \quad \quad \quad \quad \quad \quad \underline{^{k* \notin Z;}}$\\
           $\quad \quad \quad \quad \quad \quad \quad \quad \underline{^{l* \in Z}}$\\
           
            $\forall \ i \in SA | k* \in SA : s_i \leftarrow \underline{s_i - v_{k*l*}}$ \textbf{and} $Z \leftarrow Z - i$\\
            $ST \leftarrow \underline{ST + (l*,k*)}$\\
           }
     \Indm
     \caption{\emph{Shift\_activities} - \emph{Backward.}}
     \label{algo:Shift_activities_b}
\end{algorithm}

\begin{algorithm}[H]
     \SetAlgoLined
     \normalsize
     \small
     \textbf{procedure} \emph{HRS}() \quad
     $\left\{\textit{CA and SS are global structures}\right\}$\\
     \Indp
         \textcolor{blue}{$total\_HRS  = total\_HRS$} + 1\\
         $CA \leftarrow SS \leftarrow \varnothing$\\
         $SA, DC' \leftarrow$ \underline{\emph{Recursion}($n$)}\\
         {\textbf{if} $SS \neq \varnothing$ \textbf{then}\\}
         \quad  {\emph{Shift\_activities}()\\ 
         \quad   \emph{HRS}()\\} {\textbf{else} \ Report the optimal solution $DC'$}\\
    \Indm
    \caption{\emph{Hybrid Recursive Search} (HRS) - \emph{Backward.}}
    \label{algo:HRS_b}
\end{algorithm}

\subsection{Compute $v_{l*k*}$}

\begin{algorithm}[H]
     \SetAlgoNoLine
     \large
     \textbf{function} $Compute \ v_{l*k*}(Z)$\\
     \Indp
         $v_{l*k*} \leftarrow \delta$\\
         $k \ \leftarrow \varnothing$; $l \ \leftarrow \varnothing$\\
         \textbf{for} $node \in Z$ \textbf{do}\\
            \Indp
                \textbf{if} $k = \varnothing \textbf{then} \ k \leftarrow node$\\
                \textbf{for} pred $\in \ \underline{predecessors of}$ \ node \ \textbf{do}\\
                \Indp
                    $\textcolor{blue}{\emph{edge\_checked}} \leftarrow \textcolor{blue}{\emph{edge\_checked}} + 1$\\
                    \textbf{if} pred $\notin Z$ \textbf{do}\\
                        \Indp
                            
                            \eIf{$\underline{s_k - s_l < 0}$}{$current\_min = \underline{s_k - f_l - (-\delta)}$}{$current\_min = \underline{s_k - f_l}$}
                            \eIf{$current\_min < v_{l*k*}$}{
                                $v_{l*k*} \leftarrow current\_min$\\
                                 $k \leftarrow node$\\
                                 $l \leftarrow pred$\\
                                 \underline{\textbf{if} $v_{l*k*}$ < 0 \textbf{then} $v_{l*k*} = v_{l*k*} * -1$}\\
                                 }{\textbf{if} $l = \varnothing$ \textbf{then}$ \ l \leftarrow suc$\\ }
                        \Indm
                \Indm
     \Indm
     \textbf{return} ($k, l, v_{l*k*}$)\\
     
    \caption{\emph{Compute} $v_{l*k*}$ - $Backward.$}
    \label{algo:Compute_vkl_b}
\end{algorithm}


\end{document}